\documentclass[10pt,twocolumn,letterpaper]{article}

\usepackage{iccv}
\usepackage{times}
\usepackage{epsfig}
\usepackage{graphicx}
\usepackage{amsmath}
\usepackage{amssymb}
\usepackage{booktabs}
\usepackage{multicol}
\usepackage{multirow}
\usepackage{makecell}
\usepackage{comment}
\usepackage{caption}
\usepackage[dvipsnames]{xcolor}

\usepackage[pagebackref=true,breaklinks=true,letterpaper=true,colorlinks,bookmarks=false]{hyperref}

\iccvfinalcopy 

\def\iccvPaperID{10761} 

\ificcvfinal\fi

\usepackage[capitalize]{cleveref}
\crefname{section}{Sec.}{Secs.}
\Crefname{section}{Section}{Sections}
\Crefname{table}{Table}{Tables}
\crefname{table}{Tab.}{Tabs.}

\newcommand*\matr{\mathbf}

\newcommand{\Intrinsic}{\matr{K}}

\newcommand{\Rot}{\matr{R}}

\newcommand{\Line}{\mathbf{l}}
\newcommand{\RR}{\mathbb{R}}
\newcommand{\Vanishing}{\mathbf{v}}
\newcommand{\Dir}{\mathbf{d}}
\newcommand{\Basis}{\mathbf{b}}

\newcommand{\U}{\mathbf{u}}
\newcommand{\tzz}{2-0-0\emph{g}}
\newcommand{\ooz}{1-1-0\emph{g}}
\newcommand{\zoo}{0-1-1\emph{g}}

\newcommand{\todo}[1]{\textcolor{blue}{{[#1]}}}

\begin{document}

\title{Vanishing Point Estimation in Uncalibrated Images with Prior Gravity Direction}

\author{Rémi Pautrat${}^1$
\and
Shaohui Liu${}^1$
\and
Petr Hruby${}^1$
\and
Marc Pollefeys${}^{1, 2}$
\and
Daniel Barath${}^1$
\and
${}^1$ \normalsize{Department of Computer Science, ETH Zurich}
\and
${}^2$ \normalsize{Microsoft Mixed Reality and AI Zurich lab}
}

\twocolumn[{
\maketitle
\vspace{-1em}
    \centering
    \small
    \setlength{\tabcolsep}{8pt}
    \newcommand{\sz}{0.3}
    \begin{tabular}{ccc}
        \includegraphics[width=\sz\textwidth]{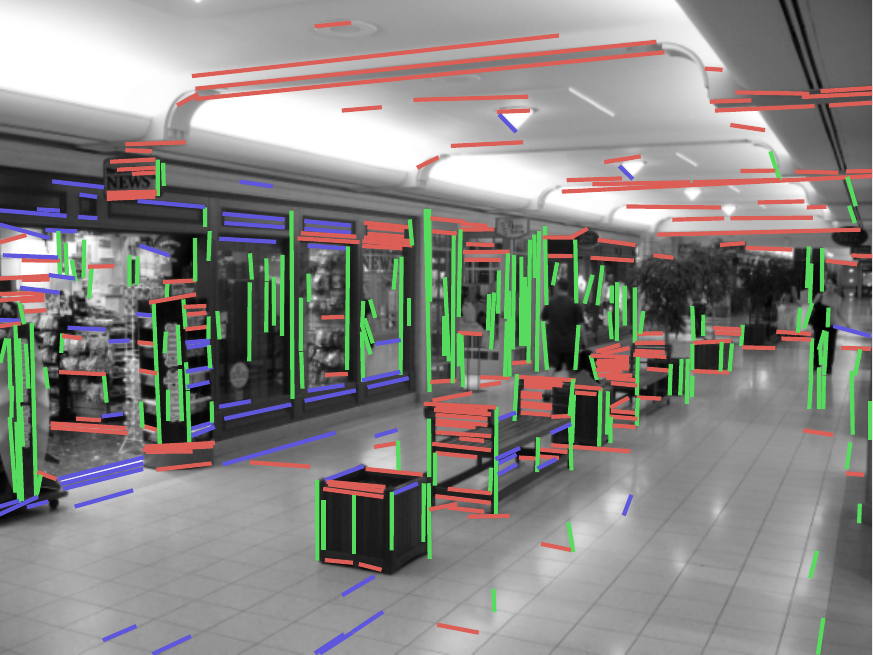} &
        \includegraphics[width=\sz\textwidth]{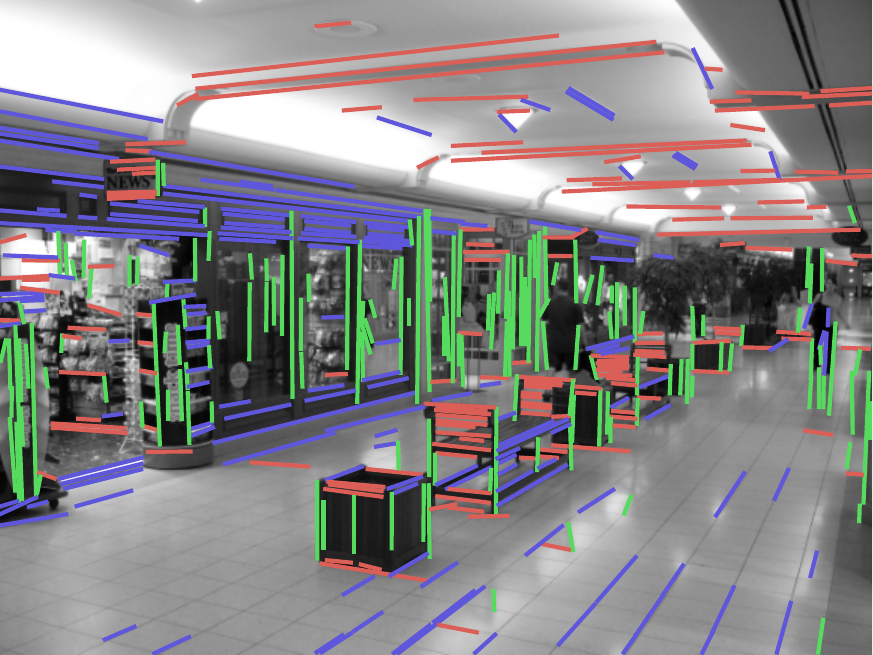} &
        \includegraphics[width=\sz\textwidth]{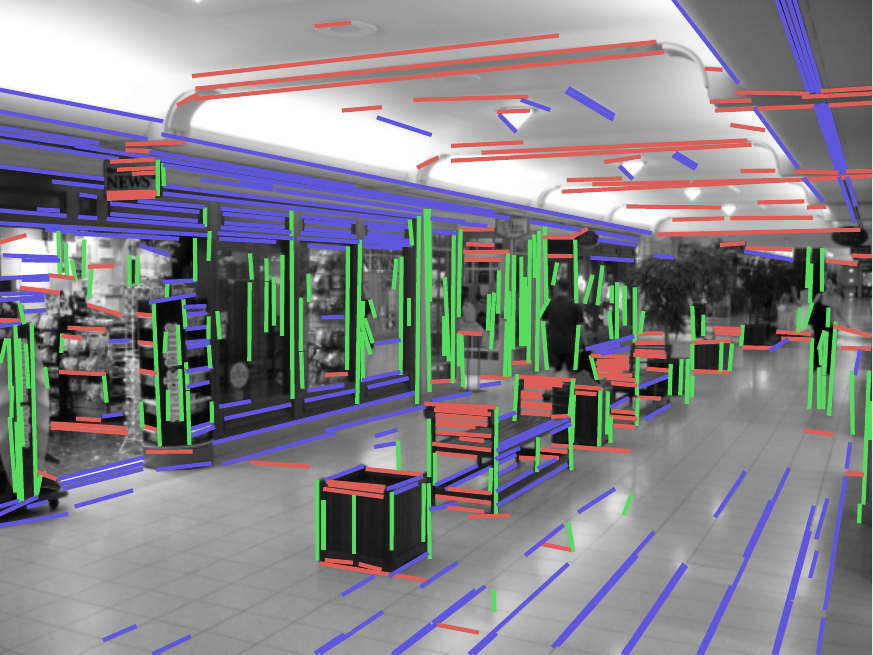} \\
        \multirow{2}{*}{\makecell{(a) 4-line solver~\cite{wildenauer2012}\\Num inliers: \textcolor{red}{\textbf{390}}, Rotation error: \textcolor{red}{\textbf{12.05$^\circ$}}}} & \multirow{2}{*}{\makecell{(b) Our 2-line solver with IMU gravity\\Num inliers: \textcolor{orange}{\textbf{455}}, Rotation error: \textcolor{orange}{\textbf{2.54$^\circ$}}}} &
        \multirow{2}{*}{\makecell{(c) Our hybrid RANSAC\\Num inliers: \textcolor{Green}{\textbf{482}}, Rotation error: \textcolor{Green}{\textbf{0.83$^\circ$}}}} \\
         & & \\[-5pt]
    \end{tabular}
    \captionof{figure}{\textbf{Vanishing point (VP) estimation with gravity prior.} We propose two new 2-line solvers for VP estimation with uncalibrated cameras leveraging a known direction. Inlier lines of RANSAC are depicted with one color per assigned VP. Compared to previous work (a), our new solvers (b) and their combination in a hybrid RANSAC (c) are more accurate.}
    \label{fig:teaser}
\vspace{15pt}
}]

\ificcvfinal\thispagestyle{empty}\fi

\begin{abstract}

We tackle the problem of estimating a Manhattan frame, i.e. three orthogonal vanishing points, and the unknown focal length of the camera, leveraging a prior vertical direction. 
The direction can come from an Inertial Measurement Unit that is a standard component of recent consumer devices, e.g., smartphones. 
We provide an exhaustive analysis of minimal line configurations and derive two new 2-line solvers, one of which does not suffer from singularities affecting existing solvers.
Additionally, we design a new non-minimal method, running on an arbitrary number of lines, to boost the performance in local optimization. 
Combining all solvers in a hybrid robust estimator, our method achieves increased accuracy even with a rough prior. 
Experiments on synthetic and real-world datasets demonstrate the superior accuracy of our method compared to the state of the art, while having comparable runtimes. 
We further demonstrate the applicability of our solvers for relative rotation estimation.
The code is available at \url{https://github.com/cvg/VP-Estimation-with-Prior-Gravity}.

\end{abstract}


\section{Introduction}
\label{sec:intro}

Projecting parallel lines of the world into a perspective image reveals a common intersection point: their vanishing point (VP).
Estimating VPs from a single image is a long-standing task in vision, as they provide critical information about the 3D structure of the scene~\cite{straub2018}.
Vanishing points are relevant in many vision applications, such as camera rotation estimation~\cite{coughlan1999,denis2008,faraz2011}, calibration~\cite{Caprile2004UsingVP}, single-view reconstruction~\cite{varsha2009}, wireframe parsing~\cite{ranade2018,Zhou_2019_ICCV}.

Vanishing point estimation methods can be categorized into two main paradigms: unconstrained VP detection discovering an undefined number of VPs~\cite{barnard1983,Quan1989DeterminingPS,zhai2016,kluger2020consac}, and approaches that make the Manhattan world assumption~\cite{coughlan1999,Bazin12,zhang2015,Li_2019_ICCV}, \ie, there are three dominant and orthogonal directions in the world. 
We opt in this paper for the second option, assuming a Manhattan world. 
This scenario is particularly relevant in human-made environments, where most applications leveraging VPs take place. 
By making this assumption, we can directly estimate the three orthogonal directions, significantly reducing the search space and simplifying the robust estimation problem.
Although it is possible to detect an unconstrained number of VPs and later orthogonalize them~\cite{denis2008}, directly looking for a set of 3 orthogonal VPs is more efficient and accurate in practice~\cite{Bazin12}.

Vanishing point detection has been extensively studied in the past. 
We distinguish five main directions of research: exhaustive search methods~\cite{magee1984,rother2002,Bazin2012RotationEA,lu2017,Qian2022ARO}, expectation-maximization-based~\cite{antone2000,denis2008}, optimization-based~\cite{bazin_cvpr_2012,Tretyak2011GeometricIP}, deep learning-based~\cite{Antunes_2017_CVPR,zhou2019neurvps,Liu_2021_ICCV,Tong_2022_CVPR} and RANSAC-based ones~\cite{Aguileraa2005ANM,Wildenauer2007VanishingPD,wildenauer2012,Bazin12,zhang2015}. 
Due to their robustness to outliers, we adopt the strand of research based on RANSAC in this paper~\cite{fischler1987}. 
To make RANSAC fast and accurate, minimal solvers are recommended in the model estimation step.
For the Manhattan world assumption, the minimum number of lines to estimate the 3 VPs from a calibrated image is 3~\cite{Bazin12}.
In the uncalibrated case with unknown focal lengths and known principal point, 4 lines are needed to recover the 3 VPs and the focal length~\cite{wildenauer2012,zhang2015}.
Assuming the principal point to be in the center of the image is a common practice~\cite{wildenauer2012,zhang2015,Qian2022ARO}, which leads to accurate focal lengths~\cite{Kanatani2005StatisticalOF}.

In this paper, we focus on estimating three orthogonal VPs in an uncalibrated image, given a known direction.
Specifically, we consider the case of a vertical prior (\ie, gravity direction) but the same theory applies to other common directions, such as the horizon line. 
Nowadays, most commercial devices, including smartphones and AR glasses, come with built-in Inertial Measurement Units (IMU) measuring the gravity direction with good precision.
Moreover, people tend to take pictures upright, providing a prior that vertical lines should roughly align with the gravity direction.
This known direction is a rich signal for VP detection, as it gives one VP in the calibrated setting~\cite{Bazin12}.

Assuming unknown focal lengths and known vertical direction is a practical problem with numerous real-world applications.
For example, it is crucial for camera systems that are re-calibrating on-the-fly or that allow zooming in and out.
An application for this task is autonomous driving, where the camera always stays upright, but its calibration may drift after several months of use, requiring an online calibration.
Furthermore, tourist pictures found on social media are often taken roughly upright, without any available camera calibration.
Finally, vanishing points provide strong constraints on the 3D rotation when calculating the relative pose of an image pair. This is particularly useful when the cameras undergo pure rotation, where traditional fundamental matrix-based solutions fail~\cite{hartley_zisserman_2004}.

We propose two new minimal solvers that address line configurations not discussed in prior work~\cite{lobo2003}.
The solvers estimate the camera rotation and focal length from a pair of lines.
Importantly, one of the proposed solvers is not affected by a singularity that makes other methods unstable when dealing with perfectly upright images.
To enable estimation from any number of lines, we also present a new non-minimal solver that can be used for local optimization in RANSAC.
Moreover, we combine all the proposed and existing solvers in a hybrid RANSAC~\cite{Camposeco2018CPVR} framework, illustrated in Figure~\ref{fig:teaser}, that adaptively decides when to use the gravity prior.
In summary, our contributions are as follows:
\begin{itemize}
    \setlength\itemsep{-1mm}
    \item We provide \textit{all line configurations leading to minimal problems} for VP estimation given known gravity but unknown focal length.
    \item We derive \textit{solutions for configurations not previously discussed in the literature}, one of which overcomes a common singularity of previous solvers.
    \item We propose \textit{a new non-minimal solver} to boost performance in the local optimization of RANSAC.
    \item We combine all the existing configurations into a hybrid RANSAC framework to \textit{increase robustness to situations with only a rough gravity prior}.
\end{itemize}
The proposed methods are evaluated on large-scale and real-world datasets, demonstrating superior accuracy compared to the state of the art when the gravity direction is known.
When we are only given a rough prior, the hybrid solution efficiently uses all solvers together to obtain results more accurately than by previous solvers. 
Furthermore, we demonstrate that such an approach is highly beneficial for relative rotation estimation in a pair of cameras.

\section{Uncalibrated Vanishing Point Estimation}

In this work, we make the Manhattan world assumption.
We are looking for 3 orthogonal vanishing points from a single image with a pinhole camera model. 
While the principal point is assumed to be in the center of the image, we consider the focal length  unknown, but the gravity direction to be known (\eg, from an IMU) or we have a rough prior on it (\eg, the image is upright). 
In this section, we first discuss existing work~\cite{lobo2003} (\ref{sec:200}) and introduce two new solvers (\ref{sec:011} and \ref{sec:110}). 
Then, we describe a new non-minimal solver in Section~\ref{sec:NMS}. 
Finally, a hybrid RANSAC approach is designed to combine all solvers in Section~\ref{sec:hybrid_ransac}.

\vspace{1mm}
\noindent
\textbf{Theoretical Background. }
A vanishing point is an intersection of 2D projections of parallel 3D lines. 
The homogeneous coordinates of the VP in the camera are
\begin{equation}
    \Vanishing_i = \Intrinsic \Rot \Dir_i, \label{eq:vp_definition}
\end{equation}
where $\Intrinsic \in \RR^3$ is the matrix of camera intrinsics, $\Rot \in \text{SO}(3)$ is the rotation of the camera, and $\Dir_i \in \RR^3$ is the direction of the line in 3D in the world coordinate system.

In this paper, we assume known principal point and square pixels. The intrinsic matrix can be simplified as
\begin{equation}
    \Intrinsic = \text{diag}(f, f, 1) \in \mathbb{R}^{3\times3},
    \label{eq:intrinsic}
\end{equation}
where $f \in \mathbb{R}$ is the focal length.
We assume that the directions $\Dir_1$, $\Dir_2$, $\Dir_3$ are mutually orthogonal, \ie, we are in a Manhattan world \cite{coughlan1999}. Directions $\Dir'_1$, $\Dir'_2$, $\Dir'_3$ in the camera coordinate system are given by $\Dir'_i = \Rot \Dir_i$, $i \in \{1,2,3\}$.

If we set the basis of the world coordinate system to be equal to directions $\Dir_1$, $\Dir_2$, $\Dir_3$ (\ie, $[\Dir_1 \ \Dir_2 \ \Dir_3] = I$), then the rotation $\Rot$ between the world and the camera coordinate system equals to $\Rot = [\Dir'_1 \ \Dir'_2 \ \Dir'_3]$.

\begin{figure}
    \centering
    \setlength{\tabcolsep}{2pt}
    \newcommand{\sz}{0.19}
    \begin{tabular}{cccc@{\hskip-5pt}c}
        \includegraphics[width=\sz\columnwidth]{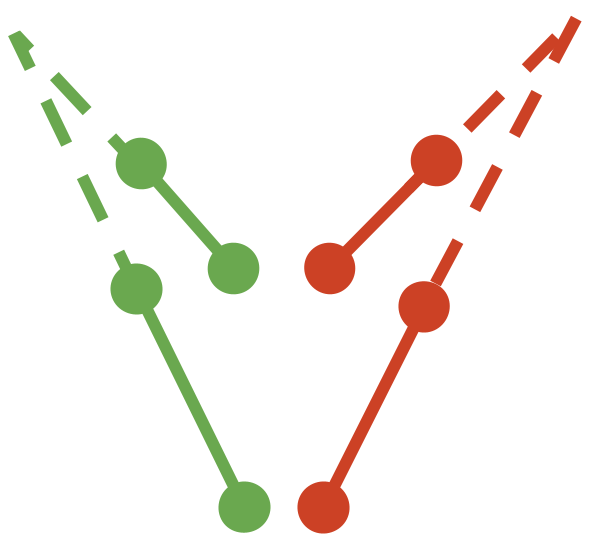} & \includegraphics[width=\sz\columnwidth]{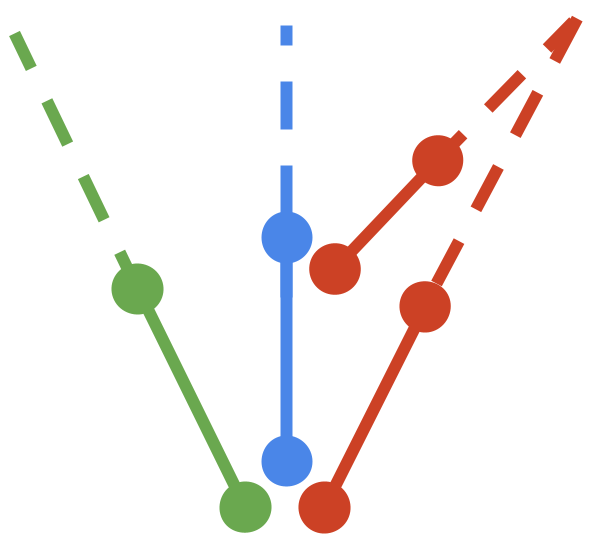} &
        \includegraphics[width=\sz\columnwidth]{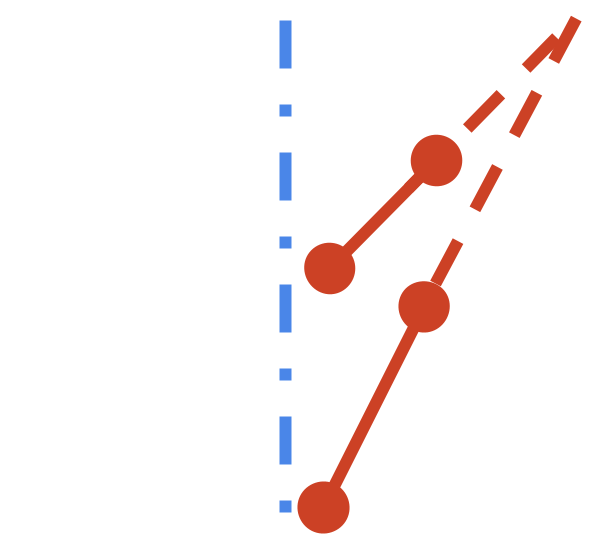} & \includegraphics[width=\sz\columnwidth]{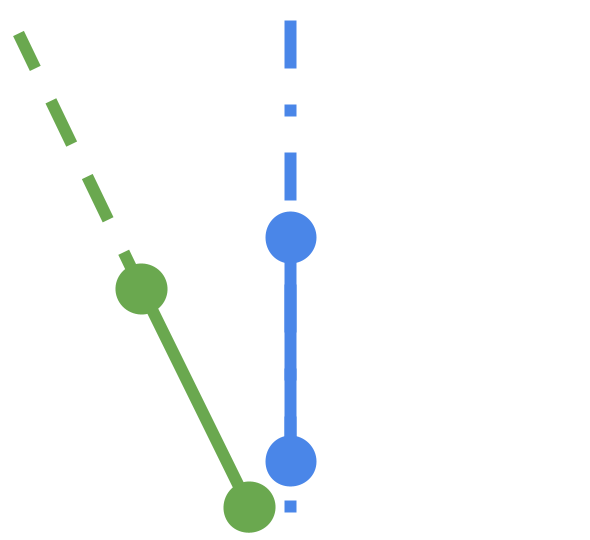} & \includegraphics[width=\sz\columnwidth]{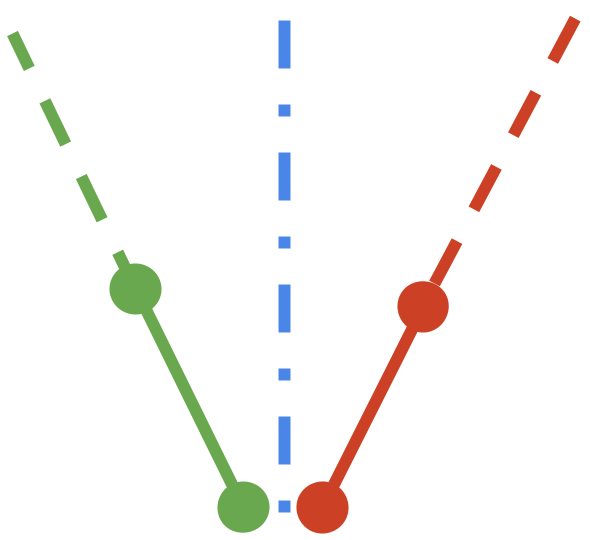} \\
        2-2-0~\cite{wildenauer2012} & 2-1-1~\cite{wildenauer2012} & \tzz~\cite{lobo2003} & \zoo & \ooz
    \end{tabular}
    \caption{\textbf{Minimal solvers in the uncalibrated case.} 
    The first two drawings display the line configuration of the 4-line solvers~\cite{wildenauer2012}, where colors correspond to VPs. The last three show the existing \cite{lobo2003} and proposed 2-line solvers with known gravity (in dashed blue). The notation $x_1\mathrm{-}x_2\mathrm{-}x_3$ means $x_i$ lines are associated with VP $i$.}
    \label{fig:solvers_overview}
\end{figure}

\subsection{Uncalibrated \tzz\ Solver~\cite{lobo2003}}
\label{sec:200}

In this section, we recap the minimal solver from \cite{lobo2003}, estimating the three orthogonal vanishing points from two lines $\Line_1$, $\Line_2$ both being associated with the same VP (third configuration in Figure~\ref{fig:solvers_overview}).
In this case, lines $\Line_1$, $\Line_2$ are projections of parallel lines in direction $\Dir'_2$. 
We further assume that the vertical direction $\Dir'_1$ is known.
We calculate vanishing point $\Vanishing_2$ as the cross product of the two lines as 
    $\Vanishing_2 = \Line_1 \times \Line_2$.
We can express the second direction as $\Dir'_2 = \Intrinsic^{-1} \Vanishing_2$ and use the constraint that $\Dir'_1$
 and $\Dir'_2$ are orthogonal as follows:
\begin{equation}
    {\Dir'_1}^\text{T} \Dir'_2 = {\Dir'_1}^\text{T} \Intrinsic^{-1} \Vanishing_2 = 0. \label{eq:orthogonality_20}
\end{equation}
Matrix $\Intrinsic^{-1}$ can be easily computed, yielding:
\begin{equation}
    (\Dir'_1(0) \Vanishing_2(0) + \Dir'_1(1) \Vanishing_2(1)) \frac{1}{f} + \Dir'_1(2) \Vanishing_2(2) = 0. \label{eq:focal_constraint}
\end{equation}
Finally, the focal length $f$ is obtained as:
\begin{equation}
    \label{eq:instability_2-0-0}
    f = -\frac{\Dir'_1(0) \Vanishing_2(0) + \Dir'_1(1) \Vanishing_2(1)}{\Dir'_1(2) \Vanishing_2(2)}.
\end{equation}
This equation implies two potential singularities when the denominator equals 0. The first one happens when $\Vanishing_2(2) = 0$, \ie lines $\Line_1$ and $\Line_2$ are parallel. This can be easily avoided by ignoring samples of parallel lines (at least one of the 2 remaining VPs will have lines that are not parallel).
The second singularity happens if $\Dir'_1(2) = 0$, \ie the gravity has no z component, which happens in datasets where the vertical direction is along the y-axis, for example.

After the focal length $f$ is found, we build the camera matrix $\Intrinsic$, compute $\Dir'_2 = \Intrinsic^{-1} \Vanishing_2$, and $\Dir'_3 = \Dir'_1 \times \Dir'_2$. Then, we build the rotation matrix as $\Rot = [\Dir'_1 \ \Dir'_2 \ \Dir'_3]$.

\subsection{Uncalibrated \zoo\ Solver}
\label{sec:011}
In this section, we derive a minimal solver to estimate the three orthogonal vanishing points and the focal length from a line $\Line_1$ stemming from the known vertical direction $\Dir'_1$ and a line $\Line_2$ stemming from a horizontal direction $\Dir'_2$.
This is the fourth configuration in Figure~\ref{fig:solvers_overview}.

Given \eqref{eq:vp_definition} and line $\Line_1$ stemming from the vertical direction, we have $\Line_1^\text{T} \Vanishing_1 = \Line_1^\text{T} \Intrinsic \Dir'_1 = 0$ that can be written as:
\begin{equation}
    (\Line_1(0) \Dir'_1(0) + \Line_1(1) \Dir'_1(1))f + \Line_1(2) \Dir'_1(2) = 0.
    \label{eq:011_elementwise}
\end{equation}
The focal length $f$ is then obtained as:
\begin{equation}
    \label{eq:instability_0-1-1}
    f = -\frac{\Line_1(2) \Dir'_1(2)}{\Line_1(0) \Dir'_1(0) + \Line_1(1) \Dir'_1(1)}.
\end{equation}
This solver also suffers from a singularity if the denominator is 0.
After finding the focal length $f$, we build the camera matrix $\Intrinsic$. 
The horizontal VP $\Vanishing_2$ equals to $\Vanishing_2 = \Intrinsic \Dir'_2$. 
As $\Line_2$ stems from $\Dir'_2$, $\Line_2^\text{T} \Vanishing_2 = \Line_2^\text{T} \Intrinsic \Dir'_2 = 0$ holds. 
There also holds ${\Dir'_1}^\text{T} \Dir'_2 = 0$. 
Thus, we calculate $\Dir'_2 = \Dir'_1 \times (\Intrinsic^\text{T} \Line_2)$. Then, we calculate $\Dir'_3 = \Dir'_1 \times \Dir'_2$, normalize all 3 directions, and build the rotation matrix $\Rot = [\Dir'_1 \ \Dir'_2 \ \Dir'_3]$.

\subsection{Uncalibrated \ooz\ Solver}
\label{sec:110}

In this section, we derive a minimal solver to estimate the three orthogonal vanishing points and focal length from two lines $\Line_1$, $\Line_2$, stemming from two mutually orthogonal horizontal directions $\Dir'_2$ and $\Dir'_3$, assuming a known vertical direction $\Dir'_1$. This is the fifth configuration in Figure~\ref{fig:solvers_overview}.

Since $\Dir'_2 \perp \Dir'_1$, $\Dir'_3 \perp \Dir'_1$, directions $\Dir'_2$, $\Dir'_3$ live in a 2D subspace of $\RR^3$ that is orthogonal to $\Dir'_1$. This subspace has an orthogonal basis $\{ \Basis_1, \Basis_2 \}$ such that $\Basis_1^\text{T}\Basis_1 = \Basis_2^\text{T}\Basis_2 = 1, \Basis_1^\text{T}\Basis_2 = 0$. Let $\U \in \RR^3$ be an arbitrary generic vector. Then, this basis can be estimated from $\Dir'_1$ as follows:
\begin{equation}
    \begin{split}
        \Basis'_1 = \Dir'_1 \times \U; \ \Basis_1 = \frac{1}{||\Basis'_1||} \Basis'_1, \\
        \Basis'_2 = \Dir'_1 \times \Basis'_1; \ \Basis_2 = \frac{1}{||\Basis'_2||} \Basis'_2. \\
    \end{split}
\end{equation}
Since $\Dir'_2 \perp \Dir'_3$, the directions $\Dir'_2, \Dir'_3$ can be obtained by rotating the basis elements $\Basis_1, \Basis_2$ by an unknown angle $\varphi$. They can be expressed as follows: 
\begin{equation}
\begin{split}
    \Dir'_2 = \cos \varphi \Basis_1 - \sin \varphi \Basis_2, \;
    \Dir'_3 = \sin \varphi \Basis_1 + \cos \varphi \Basis_2. \label{eq:011_orthogonality}
\end{split}
\end{equation}
Then, we can get the vanishing points as $\Vanishing_2 = \Intrinsic \Dir'_2$ and $\Vanishing_3 = \Intrinsic \Dir'_3$. The fact that $\Vanishing_2$ lies on $\Line_1$ and $\Vanishing_3$ lies on $\Line_2$ imposes the following two constraints on $f$ and $\varphi$ as
\begin{equation}
    \begin{split}
        \Line_1^\text{T} \Intrinsic \Dir'_2 = \cos \varphi \Line_1^\text{T} \Intrinsic \Basis_1 - \sin \varphi \Line_1^\text{T} \Intrinsic \Basis_2 = 0,\\
        \Line_2^\text{T} \Intrinsic \Dir'_3 = \sin \varphi \Line_2^\text{T} \Intrinsic \Basis_1 + \cos \varphi \Line_2^\text{T} \Intrinsic \Basis_2 = 0.
    \end{split}
\end{equation}
We replace the goniometric functions $\cos \varphi$, $\sin \varphi$ by $\frac{1-t^2}{1+t^2}$, $\frac{2t}{1+t^2}$, and multiply both equations by $1+t^2$ to obtain an equivalent system of polynomial equations as
\begin{equation}
    \begin{split}
        (1-t^2) \Line_1^\text{T} \Intrinsic \Basis_1 - 2t \Line_1^\text{T} \Intrinsic \Basis_2 = 0, \\
        2t \Line_2^\text{T} \Intrinsic \Basis_1 + (1-t^2) \Line_2^\text{T} \Intrinsic \Basis_2 = 0. \label{eq:011_constraints}
    \end{split}
\end{equation}
System \eqref{eq:011_constraints} is linear in $f$ and quadratic in $t$. 
This gives us two ways to eliminate variables. 
In the proposed solver, we first eliminate $t$, and then solve for $f$, which is shown in the following paragraph. 
An alternative approach is to eliminate the equations in reverse order, which we derive in the supplementary material. The first option performed better in our experiments, so we only use this one in the following. After we obtain $f$ and $t$, we build $\Intrinsic$ according to \eqref{eq:intrinsic}, $\cos \varphi$ and $\sin \varphi$ as $\frac{1-t^2}{1+t^2}$, $\frac{2t}{1+t^2}$. Finally, we compute $\Dir'_2$, $\Dir'_3$ according to \eqref{eq:011_orthogonality} and $\Rot$ as $[\Dir'_1 \ \Dir'_2 \ \Dir'_3]$.

\textbf{Variable Elimination.} 
In this section, we discuss how to solve \eqref{eq:011_constraints} by first eliminating $t$ and by solving for $f$.
The equations \eqref{eq:011_constraints} are quadratic in $t$. Let us use notations
\begin{equation}
    \label{eq:notations}
    \begin{split}
        \delta_1 = \Line_1(0) \Basis_1(0) + \Line_1(1) \Basis_1(1), \ \delta_2 = \Line_1(2) \Basis_1(2),\\
        \delta_3 = \Line_1(0) \Basis_2(0) + \Line_1(1) \Basis_2(1), \ \delta_4 = \Line_1(2) \Basis_2(2),\\
        \delta_5 = \Line_2(0) \Basis_2(0) + \Line_2(1) \Basis_2(1), \ \delta_6 = \Line_2(2) \Basis_2(2),\\
        \delta_7 = \Line_2(0) \Basis_1(0) + \Line_2(1) \Basis_1(1), \ \delta_8 = \Line_2(2) \Basis_1(2).
    \end{split}
\end{equation}
Then, the equations \eqref{eq:011_constraints} can be written as
\begin{equation}
\label{eq:rewriting_with_delta}
\begin{split}
    (1-t^2) (f \delta_1 + \delta_2) - 2 t (f \delta_3 + \delta_4) = 0,\\
    (1-t^2) (f \delta_5 + \delta_6) + 2 t (f \delta_7 + \delta_8) = 0.    
\end{split}
\end{equation}
Multiplying the first equation by $f \delta_5 + \delta_6$ and the second one by $f \delta_1 + \delta_2$, then subtracting them, we obtain
\begin{equation}
    \label{eq:nul_product}
    2t [ (f \delta_1 + \delta_2) (f \delta_7 + \delta_8) + (f \delta_3 + \delta_4) (f \delta_5 + \delta_6) ] = 0.
\end{equation}
If $t \neq 0$, then the second term must be $0$. If $t = 0$, \eqref{eq:rewriting_with_delta} yields $f \delta_1 + \delta_2 = f \delta_5 + \delta_6 = 0$, and so the second term is also $0$. Thus, we have in both cases:
\begin{equation}
    (f \delta_1 + \delta_2) (f \delta_7 + \delta_8) + (f \delta_3 + \delta_4) (f \delta_5 + \delta_6) = 0. \label{eq:011_quadratic}
\end{equation}
This is a univariate quadratic equation, so we obtain its roots by a closed-form formula. We only keep the positive values of $f$ and finally obtain $t$ as common roots of \eqref{eq:011_constraints}.

\subsection{Non-Minimal Solver}\label{sec:NMS}

Here, we are going to discuss how to compute vanishing points corresponding to $3$ directions from a larger-than-minimal sample. 
This step is essential to obtain accurate results, \eg, when refitting on all inliers after RANSAC finishes or in the local optimization procedure.

Suppose that we are given three vanishing points $\mathbf{v}_1$, $\mathbf{v}_2$, $\mathbf{v}_3$, either from the previously introduced minimal solvers, or from another minimal solver that does not require known gravity. 
Let $\mathcal{L}_1$, $\mathcal{L}_2$, $\mathcal{L}_3$ be three sets of lines, where each set $\mathcal{L}_i, i \in \{1,2,3
\}$ contains $n_i$ inliers of vanishing point $\mathbf{v}_i$.
First, we re-estimate each vanishing point $\Vanishing_i, i\in \{1,2,3\}$ from its inliers $\mathcal{L}_i$ using the least squares (LSQ) method (we give the exact derivation in the supplementary material).

Next, we estimate the focal length $f$ using these updated vanishing points. For every pair of vanishing points $\Vanishing_i$, $\Vanishing_j$, and \eqref{eq:vp_definition} and \eqref{eq:orthogonality_20}, 
    $\Dir_i^\text{T} \Dir_j = \Vanishing_i^\text{T} (\Intrinsic^{-1})^\text{T} \Intrinsic^{-1} \Vanishing_j = 0$
holds. This constraint can be rewritten as:
\begin{equation}
    -\Vanishing_i(2)\Vanishing_j(2) f^2 = \Vanishing_i(0)\Vanishing_j(0) + \Vanishing_i(1)\Vanishing_j(1),
\end{equation}
where the numbers in the parentheses refer to the $x$, $y$, and $w$ coordinates of the homogeneous points.
Taking $(i,j) \in \{ (1,2), (1,3), (2,3) \}$ gives three independent constraints that all are linear in $f^2$. We use the LSQ method to find $f^2$ (see the supplementary material for the full derivation). Since $f$ can not be negative, we take $f = +\sqrt{f^2}$.

Finally, we correct the vanishing points to be orthogonal. 
We compute the calibrated vanishing points $\Dir_i, i \in \{1,2,3\}$ as $\Dir_i = \matr K^{-1} \Vanishing_i$, build a matrix $\matr{D} = [\Dir_1 \ \Dir_2 \ \Dir_3]$, and decompose as $\matr{D} = \matr{U}\matr{S}\matr{V}^\text{T}$. We take the rows of matrix $\Rot = \matr{U}\matr{V}^\text{T}$ as the refined calibrated vanishing points.
Note that the VPs are further refined with a least square optimization described in \cref{sec:lo}.

\subsection{Hybrid Robust Estimation}
\label{sec:hybrid_ransac}

At this point, we have five minimal solvers for uncalibrated vanishing point estimation: two 4-line solvers without prior \cite{wildenauer2012} and three solvers with known gravity direction. 
As the best solver for each problem depends on the accuracy of the prior gravity, we employ all the solvers in a hybrid RANSAC framework to combine the advantages of both worlds. 
At each iteration of RANSAC, we first sample a minimal solver with respect to a probability distribution, which is computed with the prior distribution and the inlier ratio $\epsilon$. 
Specifically, we multiply the corresponding prior probability by $\epsilon^2$ for the 2-line solvers with gravity direction and $\epsilon^4$ for the 4-line solvers without prior, followed by normalization with the sum of the five probabilities. 
Then, similarly as in RANSAC, we run the solver on a randomly sampled minimal set, score the models, and apply our proposed local optimization when a new best model is found. The termination criterion is adaptively determined for each solver as in \cite{Camposeco2018CPVR}, depending on the inlier ratio and the predefined confidence parameter. In our experiment, we set the prior probability to be uniform across the five solvers. 

\section{Experiments}

\subsection{Synthetic Experiments}

In order to test our solvers in a fully controlled synthetic environment, we generate instances of the problem we wish to solve, together with the ground truth solution.

\begin{figure}
\centering
\includegraphics[width=0.65\linewidth,trim={3mm 0 0 0},clip]{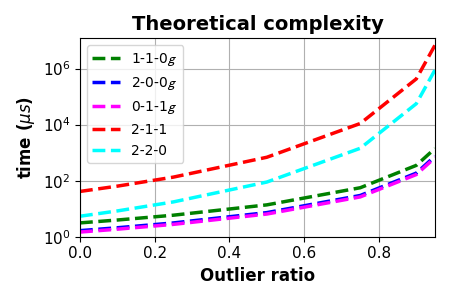}
\caption{\textbf{Theoretical RANSAC run-time}, in $\mu$s, for solvers 
2-2-0, 2-1-1~\cite{wildenauer2012}, \tzz~\cite{lobo2003}, and the proposed \ooz, and \zoo, calculated by multiplying the solver time by the theoretical number of RANSAC iterations at $0.99$ conf.}
\label{fig:complexity}
\end{figure}

\vspace{1mm}\noindent\textbf{Solver times.}
The theoretical run-times of the solvers inside RANSAC are plotted as a function of the outlier ratio in Fig.~\ref{fig:complexity}.
It is calculated as the actual time of each solver, implemented in C++, multiplied by the theoretical number of iterations RANSAC has to perform on a particular outlier level to achieve $0.99$ confidence.
The average wall-clock times of the solvers are $0.09\mu$s (2-2-0), $0.69\mu$s (2-1-1), $0.09\mu$s (\tzz), $0.08\mu$s (\zoo), and $0.17\mu$s (\ooz).
The proposed solvers are significantly faster inside RANSAC, mainly because they require fewer iterations.

\vspace{1mm}\noindent\textbf{Numerical stability.}
We generate a random rotation matrix $\Rot_\text{gt} = [\Dir'_1, \Dir'_2, \Dir'_3]$, and a random focal length $f_\text{gt}$ from a uniform distribution between $100$ and $2000$ pixels. 
We take the vanishing points $\Vanishing_1$, $\Vanishing_2$, $\Vanishing_3$ as the columns of matrix $\Intrinsic_\text{gt} \Rot_\text{gt}$.
The vertical direction is the first column $\Dir'_1$ of $\Rot_\text{gt}$. 
To generate a 3D line in direction $i$, we sample the first endpoint $\matr{X}_A$ from the Gaussian distribution with mean $[0 \ 0 \ 5]^\text{T}$ and standard deviation set to $1$. 
We sample a parameter $\lambda \in \RR$ from the normalized Gaussian distribution, and take the second endpoint as $\matr{X}_B = \matr{X}_A + \lambda \Dir'_i$. 
Then, we project both endpoints as $\matr{x}_A = \Intrinsic_\text{gt} \matr{X}_A$, $\matr{x}_B = \Intrinsic_\text{gt} \matr{X}_B$, and obtain the homogeneous coordinate of the line as $\Line = \matr{x}_A \times \matr{x}_B$.

\begin{figure}
    \centering
    \includegraphics[width=1\linewidth]{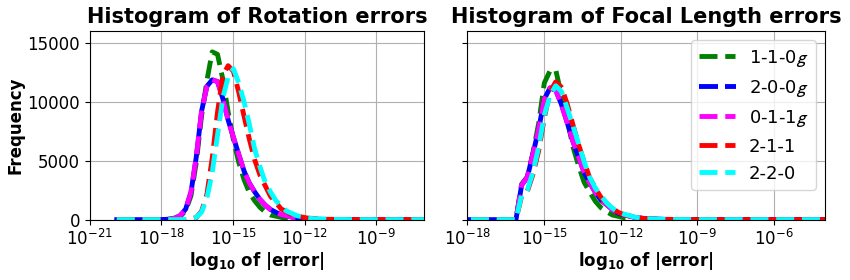}
    \caption{
    \textbf{Stability study.} 
    The frequencies ($100$k runs) of the $\text{log}_{10}$ errors in rotation and focal length estimated by solvers 2-2-0, 2-1-1~\cite{wildenauer2012}, \tzz~\cite{lobo2003}, \ooz, and \zoo.}
    \label{fig:stability_tests}
\end{figure}

Let $\Rot_\text{est}$, $f_\text{est}$ be the rotation and focal length estimated by the minimal solver. We compute the rotation error as the angle of the rotation represented as ${\Rot_\text{est}}^\text{T} \Rot_\text{gt}$, and the focal length error as $\lvert f_\text{est} - f_\text{gt} \rvert$. 
We generated $n=100000$ random problem instances and ran the solvers on the noiseless samples.
Figure \ref{fig:stability_tests} shows histograms of $\log_{10}$ rotation and focal length errors. 
The plots show that all our solvers are very stable -- there is no peak close to $10^0$.

\vspace{1mm}\noindent\textbf{Tests with noise.}
To evaluate the robustness w.r.t.\ the noise in the input, we generate minimal problems similarly as in the previous section. 
We perturb the gravity direction and the endpoints using Gaussian noise with standard deviation $\sigma_i$
To perturb the gravity direction $\Dir'_1$, we sample axis $\matr{a}_g$ uniformly from a unit sphere, and an angle $\alpha_g$ from a Gaussian distribution with zero mean and standard deviation $\sigma_g$. 
We build a rotation matrix $\Rot_g$ with angle $\alpha_g$ and axis $\matr{a}_g$, and get the noisy gravity direction as $\Rot_g \Dir'_1$. 

\begin{figure}
    \centering
    \setlength\tabcolsep{0pt}
    \setlength\extrarowheight{-3pt}
     \renewcommand{\arraystretch}{0}
     \includegraphics[width=0.9\linewidth]{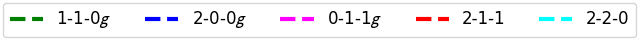}
    \begin{tabular}{c c}

       \includegraphics[width=0.5\linewidth]{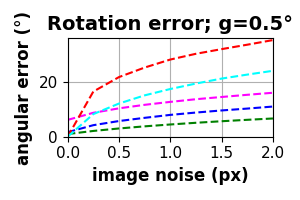}
       &
       \includegraphics[width=0.5\linewidth]{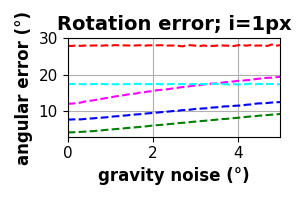}
       \\

       \includegraphics[width=0.5\linewidth]{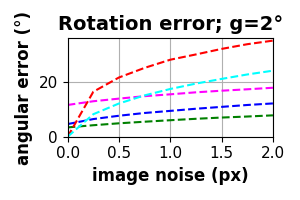}
       &
       \includegraphics[width=0.5\linewidth]{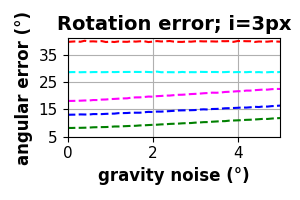}
       \\
    \end{tabular}
    
    \caption{\textbf{Noise study.} 
    The avg.\ ($100$k runs) rotation error of solvers 2-1-1~\cite{wildenauer2012}, 2-2-0~\cite{wildenauer2012}, \tzz~\cite{lobo2003}, \ooz, and \zoo\, as a function of the image (left) and gravity noise (right). 
    The noise std.\ is in the title (g -- gravity, i -- image). }
    \label{fig:noise_tests}
\end{figure}

Figure \ref{fig:noise_tests} shows the average rotation and focal length errors on different levels of $\sigma_i$ and $\sigma_g$.
In the plots on the left, the proposed solvers are significantly less sensitive to the image noise than the 4-line ones. 
In the right plots, they are more accurate than the 4-line solvers, even with some noise on the gravity prior. 
Note that accelerometers used in cars and modern smartphones have noise levels around $0.06^\circ$ (or expensive “good” ones have less than $0.02^\circ$) \cite{ding2021globally}. 
Additional experiments are shown in the supp.\ material.

\subsection{Real-World Experiments Setup}

\vspace{1mm}\noindent\textbf{Datasets.} 
We evaluate our methods on two real-world datasets. The first one is the York Urban DB dataset~\cite{yorkurban}, a classical benchmark for vanishing point estimation. It consists of 102 images of indoor and outdoor scenes, annotated with 3 orthogonal VPs. We split it into 51 images for tuning hyper-parameters and 51 for testing. When using our solvers with known gravity, we consider two cases: the gravity is given by the ground truth vertical VP and is, consequently, very accurate (we label it as \textit{IMU}), or we only use a prior gravity assuming that the images are upright (denoted as \textit{Prior}). 
In the latter case, the average gravity error w.r.t.\ the ground truth is of 3.9 degrees, showing that most pictures are indeed close to being upright.

The second dataset that we consider is ScanNet~\cite{dai2017scannet}, a large-scale indoor RGB-D dataset. We follow the evaluation of NeurVPS~\cite{zhou2019neurvps}, by finding the three main orthogonal directions aligning with the most surface normals in each image. For each of the 100 scenes of the official ScanNet validation set, we randomly sample 10 images for testing and one for validation, resulting in a test set of 1000 images and a validation set of 100 images to tune the hyper-parameters. We again choose the \textit{IMU} gravity to be the VP closest to the vertical direction, and use the prior that images should be upright. In this case, the average error of the prior w.r.t.\ the GT gravity is much larger, with 24.8$^\circ$ error.

\vspace{1mm}\noindent\textbf{Metrics.} Since the 3 detected VPs are orthogonal, we can directly derive the corresponding camera rotation. Thus, we report several metrics for rotation estimation and vanishing point detection. The rotation error is the distance between the quaternions of the GT rotation $q_\text{gt}$ and the predicted quaternions $q$ in degrees: $R_{err} = \arccos (2 (q \cdot q_\text{gt})^2 - 1)$. The rotation AUC measures the area under the recall curve at 5 / 10 / 20 degrees. The VP error is the average of the angular errors in degrees between the predicted VPs and the GT ones. Finally, we regularly sample 20 thresholds between 0 and 10 degrees, plot the curve of the percentage of images successfully recovering VPs that are close enough to the GT VPs up to these thresholds, and compute the Area Under the Curve to get the VP AUC, as in \cite{kluger2020consac}.

\begin{figure}
    \centering
    \scriptsize
    \newcommand{\val}{0.32}
    \setlength{\tabcolsep}{1pt}
    \begin{tabular}{ccc}
        2-2-0 solver~\cite{wildenauer2012} & Hybrid w/ prior gravity & Hybrid w/ GT gravity \\
        \includegraphics[width=\val\columnwidth]{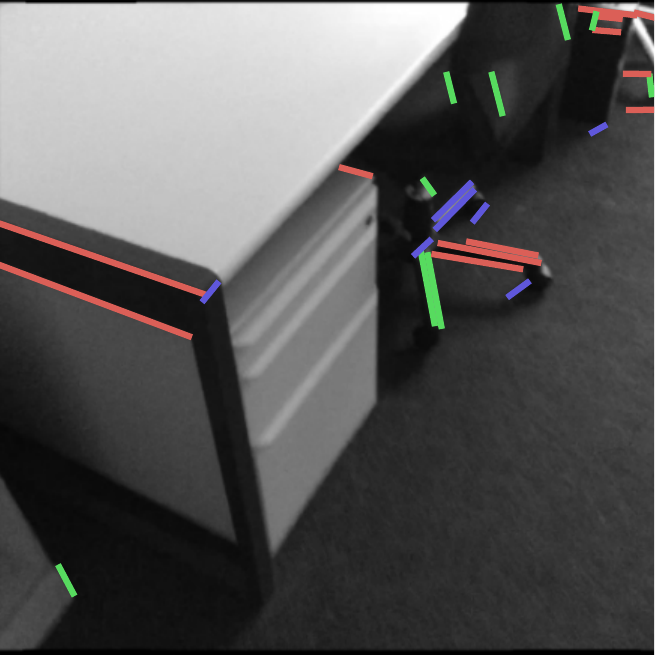} & \includegraphics[width=\val\columnwidth]{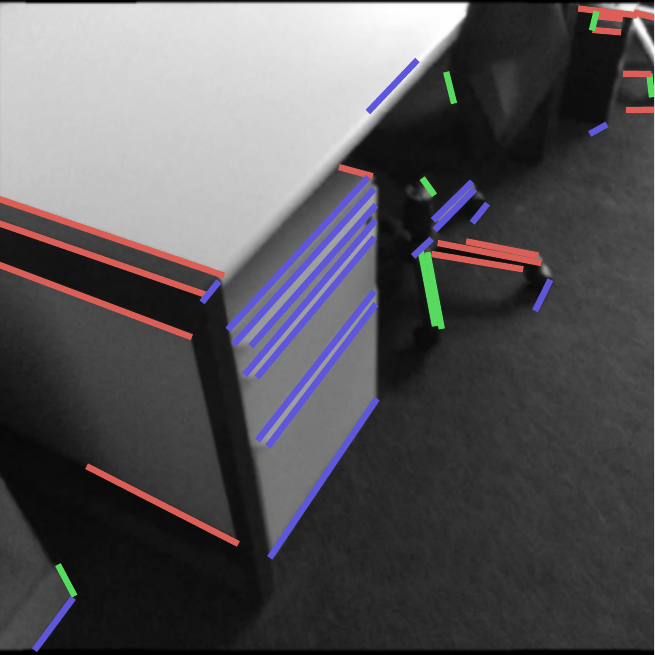} & \includegraphics[width=\val\columnwidth]{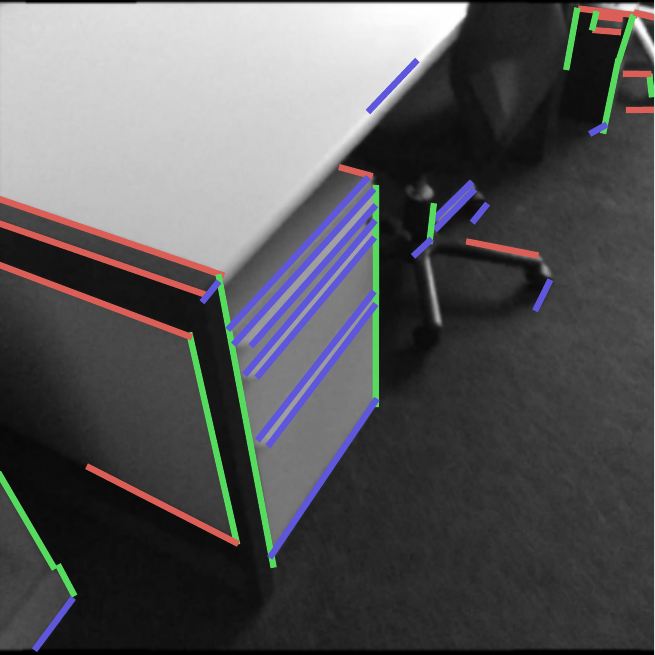} \\
        \includegraphics[width=\val\columnwidth]{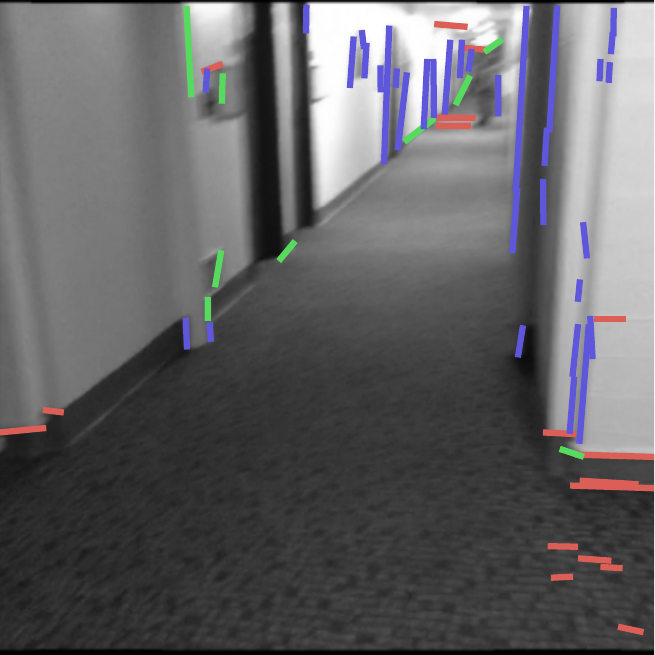} & \includegraphics[width=\val\columnwidth]{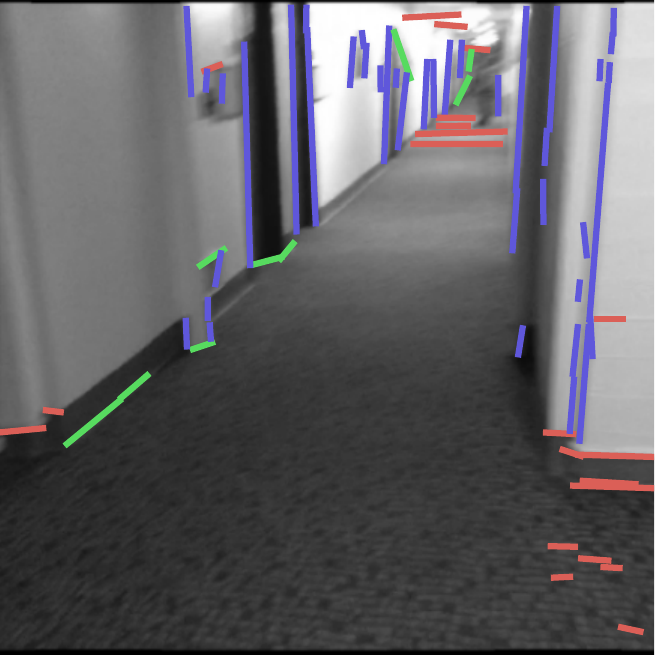} & \includegraphics[width=\val\columnwidth]{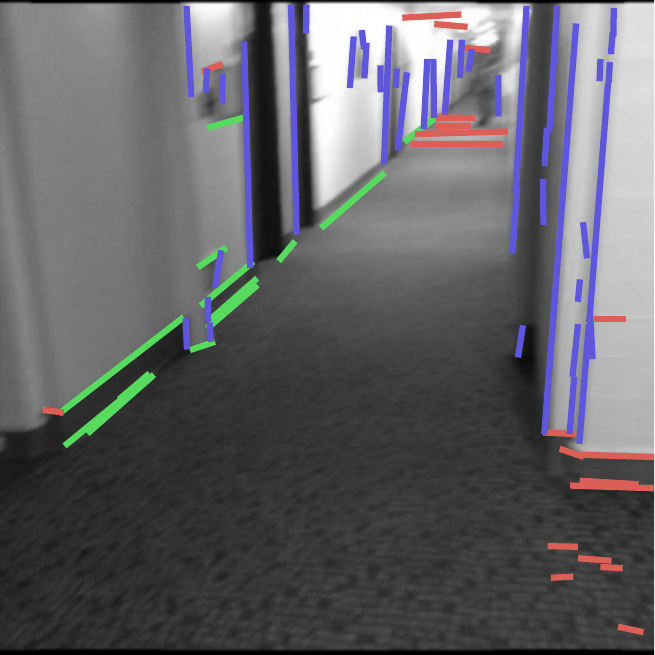} \\
    \end{tabular}
    \caption{\textbf{Visual examples.} We provide examples of VP estimation on ScanNet~\cite{dai2017scannet} by the 2-2-0 solver~\cite{wildenauer2012}, our hybrid RANSAC with prior gravity, and with GT gravity. Lines are colored according to their associated VP. Even with a rough prior, our hybrid solver obtains more accurate results than the baseline.}
    \label{fig:examples}
\end{figure}

\vspace{1mm}\noindent\textbf{Baselines.} For all following experiments, we apply the widely used Line Segment Detector~\cite{von2008lsd} to extract line segments from images. All RANSAC methods are run with a minimum of 1000 iterations and we report the median results after 30 runs on York Urban and 10 runs on ScanNet. We compare our minimal solvers with the two previously existing 4-line solvers for uncalibrated cameras~\cite{wildenauer2012}: 2-2-0 and 2-1-1 (first two configurations in Figure~\ref{fig:solvers_overview}), and the 2-line solver \tzz\ of \cite{lobo2003}. We show the individual results of our two new 2-line solvers \ooz\ and \zoo, as well as of our hybrid RANSAC combining all five solvers. Figure ~\ref{fig:examples} illustrates the extracted vanishing points of different methods on the challenging ScanNet dataset~\cite{dai2017scannet}.

\subsection{Impact of Local Optimization}
\label{sec:lo}

We start by evaluating the performance gain from our proposed non-minimal solver and the resulting local optimization (LO) applied in RANSAC whenever a new best model is found. 
We follow the LO-RANSAC implementation of \cite{Lebeda2012loransac}, by first extracting inliers of the best model returned by one of the considered minimal solvers and then running 100 iterations of LO on random subsets of these inliers. Each LO step generates a new model from the non-minimal solver, refines it with a least squares (LS) optimization fitting the VP to the inlier lines, and stores the new model if it improves the previous best model. 
The LS optimization minimizes the sum of squared distances between each VP and its corresponding inlier lines, and is implemented in Ceres~\cite{ceres} with the Levenberg–Marquardt algorithm~\cite{lma}.
We compare our LO with the Iter method proposed in \cite{zhang2015} using the exact same pipeline as ours, but replacing the model of our non minimal solver with the best model returned by the minimal solvers instead.

We show the results on York Urban in Table~\ref{tab:yud_lo} without LO, with Iter, and our proposed LO. We only show the results with prior gravity, as it is where the LO brings the most significant contribution. 
It can be seen from the table that not using any LO yields very poor results, thus highlighting the importance of LO. 
While Iter brings a large improvement, our proposed LO obtains the best results in all metrics by a very large margin. 
This overall large boost of performance of our LO comes at the price of an increased computation time -- still close to real-time. 
It is possible to reduce the run-time by decreasing the number of LO iterations, and trading performance for efficiency.

\begin{table}[]
    \centering
    \scriptsize
    \setlength{\tabcolsep}{5pt}
    \begin{tabular}{ccccccc}
        \toprule
        \multirow{2}{*}{Solver} & \multirow{2}{*}{LO} & \multicolumn{2}{c}{Rotation estimation} & \multicolumn{2}{c}{VP estimation} & \multirow{2}{*}{\makecell{Time\\(ms)}} \\
        \cmidrule(lr){3-4} \cmidrule(lr){5-6}
         & & Err $\downarrow$ & AUC $\uparrow$ & Err $\downarrow$ & AUC $\uparrow$ &  \\
        \midrule
        \multirow{3}{*}{\makecell{2-2-0~\cite{wildenauer2012}\\No gravity}} & No & 6.04 & \phantom{1}9.7 / 36.0 / 63.8 & 4.96 & 5.35 & \textbf{65} \\
        & Iter~\cite{zhang2015} & 1.77 & 58.3 / 76.6 / 87.9 & 1.87 & 7.85 & 78 \\
        & Ours & \textbf{1.51} & \textbf{67.6 / 83.6 / 91.8} & \textbf{1.34} & \textbf{8.93} & 119 \\
        \cmidrule(lr){1-2}
        \multirow{3}{*}{\makecell{2-1-1~\cite{wildenauer2012}\\No gravity}} & No & 6.87 & \phantom{1}7.9 / 32.1 / 58.7 & 5.23 & 5.14 & \textbf{40} \\
        & Iter~\cite{zhang2015} & 2.07 & 53.0 / 74.9 / 87.3 & 1.99 & 7.57 & 57 \\
         & Ours & \textbf{1.52} & \textbf{68.3 / 83.2 / 90.6} & \textbf{1.35} & \textbf{8.88} & 103 \\
        \cmidrule(lr){1-2}
        \multirow{3}{*}{\makecell{\tzz~\cite{lobo2003}\\+ Prior}} & No & 25.09 & \phantom{1}2.3 / \phantom{1}5.0 / 12.0 & 7.63 & 2.47 & \textbf{68} \\
        & Iter~\cite{zhang2015} & 25.08 & \phantom{1}2.3 / \phantom{1}5.0 / 12.0 & 7.61 & 2.48 & 72 \\
         & Ours  & \phantom{1}\textbf{1.25} & \textbf{64.3 / 78.0 / 86.3} & \textbf{1.79} & \textbf{8.43} & 121 \\
        \cmidrule(lr){1-2}
        \multirow{3}{*}{\makecell{\zoo\\+ Prior}} & No & 25.18 & \phantom{1}2.3 / \phantom{1}5.1 / 14.1 & 7.60 & 2.68 & \textbf{40} \\
        & Iter~\cite{zhang2015} & 24.95 & \phantom{1}5.0 / \phantom{1}9.6 / 16.3 & 7.35 & 2.76 & 47 \\
         & Ours & \textbf{1.37} & \textbf{68.4 / 84.1 / 92.1} & \textbf{1.28} & \textbf{8.80} & 142 \\
        \cmidrule(lr){1-2}
        \multirow{3}{*}{\makecell{\ooz\\+ Prior}} & No & 4.48 & 26.0 / 44.2 / 59.2 & 4.79 & 5.35 & \textbf{44} \\
        & Iter~\cite{zhang2015} & 1.63 & 54.5 / 69.3 / 77.6 & 2.60 & 7.62 & 52 \\
         & Ours & \textbf{1.20} & \textbf{70.1 / 84.1 / 91.1} & \textbf{1.29} & \textbf{8.87} & 84 \\
        \cmidrule(lr){1-2}
        \multirow{3}{*}{\makecell{Hybrid\\+ Prior}} & No & 4.59 & 26.4 / 46.1 / 64.9 & 4.30 & 5.95 & \textbf{43} \\
        & Iter~\cite{zhang2015} & 1.51 & 61.7 / 79.0 / 88.5 & 1.69 & 8.19 & 52 \\
        & Ours & \textbf{1.13} & \textbf{71.5 / 85.8 / 92.9} & \textbf{1.16} & \textbf{8.97} & 100 \\
        \bottomrule
    \end{tabular}
    \caption{\textbf{Local optimization on the YorkUrban dataset~\cite{yorkurban}.} We report median errors in degrees and AUCs over 30 runs. The rotation AUC is given 5$^\circ$ / 10$^\circ$ / 20$^\circ$.}
    \label{tab:yud_lo}
\end{table}

\subsection{Comparison to Previous Works}

We compare our new solvers to the baselines on York Urban DB~\cite{yorkurban} in Table~\ref{tab:yud_uncalibrated} and on ScanNet~\cite{dai2017scannet} in Table~\ref{tab:scannet_uncalibrated}, where our LO has been applied to all methods. Several conclusions can be drawn from these two tables. First, our two new solvers significantly outperform the baselines when IMU information is available, with, for example, an improvement of up to 27\% in median rotation error over the best baselines on York Urban, and up to 40\% on ScanNet. 

Second, our solvers using the prior gravity obtain similar results as the baselines without gravity information, showing that even with a rough estimate of the gravity, it is not detrimental to use our solvers. 
This is especially true for ScanNet, where the prior gravity is particularly noisy with 24.8$^\circ$ of average error w.r.t.\ to the GT gravity. 
The better performance of the \ooz\ solver is mostly due to the fact that it is the only gravity-based solver that does not suffer from a singularity when the vector $[0, -1, 0]$ is used as gravity vector. To prevent the division by zero in Equations~\ref{eq:instability_2-0-0} and~\ref{eq:instability_0-1-1}, we perturb the prior with small random noise.

Last but not least, the hybrid approach is always at least as good as all previous solvers, and it obtains the best results when used with the prior gravity, with for instance an improvement of 12\% in rotation AUC on ScanNet. This can be explained by the fact that whenever the prior is close enough to the GT gravity, our proposed solvers will yield their best performance, while when the prior becomes too noisy, the previous 2-2-0 and 2-1-1 solvers can help improving the performance.

In brief, the proposed solvers lead to significantly better accuracy when the gravity is accurate, and are not worse than the baselines when it is inaccurate. 
Hybrid RANSAC leads to the best results when having only a rough gravity prior, \ie, the image is upright.

\begin{table}[]
    \centering
    \scriptsize
    \setlength{\tabcolsep}{5pt}
    \begin{tabular}{lcccccc}
        \toprule
        \multirow{2}{*}{Solver} & \multirow{2}{*}{Gravity} & \multicolumn{2}{c}{Rotation estimation} & \multicolumn{2}{c}{VP estimation} & \multirow{2}{*}{\makecell{Time\\(ms)}} \\
        \cmidrule(lr){3-4} \cmidrule(lr){5-6}
         & & Err $\downarrow$ & AUC $\uparrow$ & Err $\downarrow$ & AUC $\uparrow$ &  \\
        \midrule
        2-2-0~\cite{wildenauer2012} & \multirow{2}{*}{No} & 1.51 & 67.6 / 83.6 / 91.8 & 1.34 & 8.93 & 119 \\
        2-1-1~\cite{wildenauer2012} & & 1.52 & 68.3 / 83.2 / 90.6 & 1.35 & 8.88 & 103 \\
        \cmidrule(lr){1-2}
        \tzz~\cite{lobo2003} & \multirow{4}{*}{Prior} & 1.25 & 64.3 / 78.0 / 86.3 & 1.79 & 8.43 & 121 \\
        \zoo & & 1.37 & 68.4 / 84.1 / 92.1 & 1.28 & 8.80 & 142 \\
        \ooz & & 1.20 & 70.1 / 84.1 / 91.1 & 1.29 & 8.87 & 84 \\
        Hybrid & & 1.13 & 71.5 / 85.8 / 92.9 & 1.16 & 8.97 & 100 \\
        \cmidrule(lr){1-2}
        \tzz~\cite{lobo2003} & \multirow{4}{*}{IMU} & 1.13 & 71.1 / 84.7 / 92.5 & 1.22 & 8.99 & 99 \\
        \zoo & & 1.13 & 71.4 / 84.6 / 92.4 & \textbf{1.07} & 9.04 & 112 \\
        \ooz & & \textbf{1.10} & \textbf{73.2 / 86.6 / 93.3} & 1.08 & \textbf{9.12} & \textbf{66} \\
        Hybrid & & 1.12 & 72.5 / 86.3 / 93.2 & 1.11 & 9.08 & 79 \\
        \bottomrule
    \end{tabular}
    \caption{\textbf{Rotation and vanishing points estimation on the YorkUrban DB dataset~\cite{yorkurban} in the uncalibrated case.} We report median errors in degrees and AUCs over 30 runs. The rotation AUC is given at error thresholds 5$^\circ$ / 10$^\circ$ / 20$^\circ$. We consider using either no gravity information, a prior on the vertical direction (\eg the y axis should be vertical), or the ground truth vertical direction (\eg obtained through IMU). Our proposed LO is applied to all methods.}
    \label{tab:yud_uncalibrated}
\end{table}

\begin{table}[]
    \centering
    \scriptsize
    \setlength{\tabcolsep}{5pt}
    \begin{tabular}{lcccccc}
        \toprule
        \multirow{2}{*}{Solver} & \multirow{2}{*}{Gravity} & \multicolumn{2}{c}{Rotation estimation} & \multicolumn{2}{c}{VP estimation} & \multirow{2}{*}{\makecell{Time\\(ms)}} \\
        \cmidrule(lr){3-4} \cmidrule(lr){5-6}
         & & Err $\downarrow$ & AUC $\uparrow$ & Err $\downarrow$ & AUC $\uparrow$ &  \\
        \midrule
        2-2-0~\cite{wildenauer2012} & \multirow{2}{*}{No} & 43.47 & \phantom{1}6.8 / 13.8 / 21.5 & 6.50 & 3.67 & 130 \\
        2-1-1~\cite{wildenauer2012} & & 42.07 & \phantom{1}7.0 / 14.3 / 22.0 & 6.42 & 3.75 & \textbf{112} \\
        \cmidrule(lr){1-2}
        \tzz~\cite{lobo2003} & \multirow{4}{*}{Prior} & 43.99 & \phantom{1}6.5 / 13.1 / 20.2 & 6.60 & 3.55 & 131 \\
        \zoo & & 42.59 & \phantom{1}5.8 / 12.2 / 19.9 & 6.74 & 3.49 & 247 \\
        \ooz & & 43.22 & \phantom{1}6.4 / 12.9 / 20.8 & 6.64 & 3.49 & 273 \\
        Hybrid & & 42.72 & \phantom{1}7.9 / 16.2 / 24.8 & 5.89 & 4.21 & 160 \\
        \cmidrule(lr){1-2}
        \tzz~\cite{lobo2003} & \multirow{4}{*}{IMU} & 25.79 & 21.0 / 29.7 / 37.3 & 2.94 & 7.18 & 200 \\
        \zoo & & 26.67 & 19.0 / 27.7 / 35.5 & 3.23 & 6.88 & 229 \\
        \ooz & & \textbf{24.99} & \textbf{21.4 / 30.3 / 38.1} & \textbf{2.87} & \textbf{7.25} & 168 \\
        Hybrid & & 25.83 & 20.3 / 28.9 / 36.6 & 3.09 & 6.94 & 190 \\
        \bottomrule
    \end{tabular}
    \caption{\textbf{Rotation and vanishing point estimation on the ScanNet dataset~\cite{dai2017scannet}.} We report median errors in degrees and AUCs over 10 runs. The rotation AUC is given at thresholds 5$^\circ$ / 10$^\circ$ / 20$^\circ$. Our proposed LO is applied to all methods. Note the higher performance of the hybrid RANSAC with prior gravity.}
    \label{tab:scannet_uncalibrated}
\end{table}

\subsection{Focal Length Estimation}

Similarly as for the previous uncalibrated solvers~\cite{wildenauer2012}, our method provides an estimate of the focal length in addition to the predicted VPs. 
We study here the quality of the prediction, by computing the relative error between the predicted and GT focal lengths. The results are reported in Table~\ref{tab:f_estimation} for the case with known gravity from IMU without LO, with Iter, and our LO. 
Our solvers and their hybrid version bring a significant improvement over the baselines without known gravity on both York Urban and ScanNet. 
Using our LO instead of Iter also generally improves the quality of the predicted focal length, in particular on York Urban. 
Thus, this experiment shows that the focal length can be extracted from VPs, and the knowledge of the gravity direction also helps with the camera calibration. 
Note that the most accurate focal lengths on ScanNet are obtained by the Iter method, but that it uses our proposed \ooz\ solver. 
Moreover, the difference between Iter and our LO is marginal, \ie, only $0.6$\%.

\begin{table}[]
    \centering
    \scriptsize
    \setlength{\tabcolsep}{5.5pt}
    \newcommand{\groupspace}{15pt}
    \begin{tabular}{l@{\hskip\groupspace}ccc@{\hskip\groupspace}ccc}
        \toprule
        \multirow{2}{*}{Solver} & \multicolumn{3}{c}{YorkUrban DB~\cite{yorkurban}} & \multicolumn{3}{c}{ScanNet~\cite{dai2017scannet}}  \\
        \cmidrule(lr{\groupspace}){2-4} \cmidrule(lr){5-7}
         & No LO & Iter~\cite{zhang2015} & Ours & No LO & Iter~\cite{zhang2015} & Ours \\
        \midrule
        2-2-0~\cite{wildenauer2012} & 0.262 & 0.104 & 0.039 & 0.527 & 0.405 & 0.252 \\
        2-1-1~\cite{wildenauer2012} & 0.176 & 0.216 & 0.041 & 0.501 & 0.394 & 0.265 \\
        \tzz~\cite{lobo2003} & 0.075 & 0.062 & 0.041 & 0.184 & 0.126 & 0.136 \\
        \zoo & 0.059 & 0.051 & 0.059 & 0.204 & 0.201 & 0.182 \\
        \ooz & 0.045 & 0.049 & \textbf{0.031} & 0.167 & \textbf{0.123} & 0.129 \\
        Hybrid & 0.065 & 0.058 & \textbf{0.031} & 0.147 & 0.132 & 0.134 \\
        \bottomrule
    \end{tabular}
    \caption{\textbf{Focal length estimation on the YorkUrban DB~\cite{yorkurban} and ScanNet~\cite{dai2017scannet} datasets.} We report the median relative distance to the ground truth focal length with and without local optimization.}
    \label{tab:f_estimation}
\end{table}

\subsection{Application: Relative Rotation with IMU}

To evaluate the solvers on a downstream application, we propose to leverage the predicted VPs and focal length to estimate the relative rotation in two uncalibrated images.
The LaMAR dataset~\cite{sarlin2022lamar} is a real-world dataset for benchmarking localization and mapping.
We considered the sequences of images coming from the validation query images of the CAB building, consisting of 4 sequences of Hololens data (images, GT poses and IMU gravity), alternating between indoors and outdoors.
Taking each pair of successive images provides a reasonable overlap for relative rotation estimation, with a total of 1415 pairs.
The Hololens is an AR headset and the images are often rotated, making the information of gravity -- provided by the IMU sensor of the headset -- essential for accurate pose estimation.
While the calibration of the Hololens is provided, a device used over several months is subject to drift and online recalibration is necessary to keep an accurate output.
Thus, we did not use the GT calibration of the dataset, and only compared methods for uncalibrated images, \ie, assuming an unknown focal length and principal point in the center of the image.

We leverage the same solvers as in previous sections to obtain sets of 3VPs and the focal length for pairs of images, independently. We then match the VPs by matching lines between the two images, checking for every pair of VPs the number of matched lines in common, and solving the optimal assignment between VPs. Each set of VPs can be concatenated in a matrix to form a camera rotation matrix, and the two matrices are combined to obtain the relative rotation between the images. We introduce here additional baselines based on keypoints matching. The first one is the 7-point algorithm for fundamental matrix estimation~\cite{hartley_zisserman_2004}, from which a focal length can be extracted~\cite{Hartley2001ExtractionOF}, allowing to convert the matrix to an essential matrix and to extract the corresponding relative rotation. We use the OpenCV implementation. The second one is a 6-point algorithm retrieving the essential matrix and a shared focal length between both images~\cite{Li2006ASS}. Finally, the 4-point algorithm retrieving the relative pose given known gravity and unknown focal length~\cite{Ding_2020_CVPR} is also compared. We use the authors' code for the last two methods. Points are detected and matched with SuperPoint~\cite{detone18superpoint} + SuperGlue~\cite{sarlin20superglue}, and lines are matched by associating line endpoints with SuperGlue.
We report the median rotation error, rotation AUC at a 5$^\circ$ / 10$^\circ$ / 20$^\circ$ error, and the relative focal length error.

Results can be found in Table~\ref{tab:lamar}. The 7-point algorithm is the de facto method used to obtain a relative pose from point correspondences from uncalibrated cameras, and scores well in this task. The 6-point solver is on the contrary more unstable and yields poorer results. Adding the information about gravity significantly boost the performance of the VP solvers, as can be seen in the last two lines. Our hybrid solver obtains the strongest performance overall, by leveraging the IMU information in three of its five solvers.
Known gravity is a strong asset here, especially in such a dataset, where pairs of images have large pure rotations. On the contrary, point-based methods such as the 7-point algorithm~\cite{hartley_zisserman_2004} are degenerate for pure rotations.

\begin{table}[]
    \centering
    \scriptsize
    \setlength{\tabcolsep}{4pt}
    \begin{tabular}{lccccc}
        \toprule
        \multirow{2}{*}{Solver} & \multirow{2}{*}{Gravity} & \multicolumn{2}{c}{Rotation estimation} & f estimation & \multirow{2}{*}{\makecell{Time\\(ms)}} \\
        \cmidrule(lr){3-4} \cmidrule(lr){5-5}
         & & Err $\downarrow$ & AUC $\uparrow$ & Err $\downarrow$ &  \\
        \midrule
        F from 7 pts~\cite{hartley_zisserman_2004,Hartley2001ExtractionOF} & \multirow{4}{*}{No} & \phantom{1}5.19 & 26.9 / 43.4 / 61.3 & 0.253 & \phantom{1}\textbf{11} \\
        E+f from 6 pts~\cite{Li2006ASS} &  & 93.65 & \phantom{1}3.7 / \phantom{1}9.2 / 17.3 & 1.000 & 459 \\
        2-2-0~\cite{wildenauer2012} &  & 12.02 & \phantom{1}2.4 / 13.6 / 36.7 & 0.287 & \phantom{1}28 \\
        2-1-1~\cite{wildenauer2012} &  & 10.06 & \phantom{1}3.7 / 18.7 / 42.8 & 0.221 & \phantom{1}38 \\
        \cmidrule(lr){1-2}
        R+f from 4 pts~\cite{Ding_2020_CVPR} & \multirow{3}{*}{IMU} & 22.53 & 16.0 / 24.3 / 33.9 & 0.869 & 834 \\
        \tzz~\cite{lobo2003} &  & \phantom{1}3.01 & 36.4 / 54.3 / 68.6 & 0.107 & \phantom{1}56 \\
        Hybrid (Ours) &  & \phantom{1}\textbf{2.89} & \textbf{38.1 / 56.8 / 71.6} & \textbf{0.093} & \phantom{1}52 \\
        \bottomrule
    \end{tabular}
    \caption{\textbf{Relative rotation and focal length estimation on the LaMAR dataset~\cite{sarlin2022lamar}.} We report the median error in degrees and AUC at error thresholds 5$^\circ$ / 10$^\circ$ / 20$^\circ$ for the rotation, as well as relative focal length error, over 5 runs. Our LO is applied to all VP solvers, and the gravity-based solvers leverage the IMU of the Hololens.}
    \label{tab:lamar}
\end{table}

\section{Conclusion}

In summary, we propose two new minimal solvers for VP estimation using 2 lines and a known gravity direction for an uncalibrated camera. 
Our solvers obtain better accuracy in VP, rotation and focal length estimations than previous works.
Furthermore, even a coarse prior on the gravity direction is enough to get comparable or better results to previous baselines. 
In order to automatically leverage the most stable solver in a data-dependent manner, we combine our two 2-line solvers with three previously existing solvers into a single hybrid robust estimator. 
We additionally introduce a new non-minimal solver significantly increasing the overall performance of all methods.

While the gravity is the most common direction that is easily accessible, our method could also be applied with other known directions, for instance the horizon line. 
There are several possible improvements that can further increase the accuracy, such as cleverly sampling the pairs of lines directly into a \tzz, \ooz, or \zoo\ configuration, extending the methods to unknown principal point, or developing a Bayesian reasoning on the noise of the prior gravity to better select solvers in our hybrid RANSAC.

{\footnotesize
\vspace{0.5em}
\noindent \textbf{Acknowledgments.}
We would like to warmly thank Linfei Pan for helping to review this paper. Daniel Barath was supported by the ETH Postdoc Fellowship.
}

\newpage

\twocolumn[
    {\centering \Large Supplementary Material \\[1ex]}
    \vspace*{3ex}
]

 \appendix

In the following, we provide additional details about our approach. Section~\ref{sec:least_square} gives the complete derivation of our proposed non minimal solver. Section~\ref{sec:derivation_details} offers additional details and derivations that were not covered in the main paper. Section~\ref{sec:synthetic_tests} displays more synthetic experiments with our proposed solvers. Section~\ref{sec:lo_supp} provides an ablation of our proposed Local Optimization (LO) on ScanNet~\cite{dai2017scannet} and with different numbers of iterations. Section~\ref{sec:magsac} offers insights about the generalization of our methods to other RANSAC strategies. Finally, Section~\ref{sec:visualizations} shows multiple visualizations of vanishing point estimation.

\section{Complete Derivations of the Non Minimal Solver}
\label{sec:least_square}

Section 2.4 of the main paper introduces our non minimal solver to estimate the orthogonal vanishing points and unknown focal length from an existing set of three vanishing points and their inlier lines. We describe here in more details the two least square methods that are used in this solver.

Re-using the same notations, we are given three vanishing points $\mathbf{v}_1$, $\mathbf{v}_2$, $\mathbf{v}_3$, and three sets of lines $\mathcal{L}_1$, $\mathcal{L}_2$, $\mathcal{L}_3$, where each set $\mathcal{L}_i, i \in \{1,2,3
\}$ contains $n_i$ inliers of vanishing point $\mathbf{v}_i$.
The first step is to re-estimate each vanishing point $\Vanishing_i, i\in \{1,2,3\}$ from its inliers $\mathcal{L}_i$ using the least squares (LSQ) method. For this, we write the sum of distances between each inlier line $\mathbf{l}_j \in \mathcal{L}_i$ and the corresponding VP $\Vanishing_i$, using homogeneous coordinates:
\begin{equation}
    \sum_{\mathbf{l}_j \in \mathcal{L}_i} d(\mathbf{l}_j, \Vanishing_i)
    = \sum_{\mathbf{l}_j \in \mathcal{L}_i} \frac{|\mathbf{l}_j^T \Vanishing_i|}{\sqrt{l_j(0)^2 + l_j(1)^2}}.
\end{equation}
Introducing the $n_i \times 3$ matrix $M_i$, defined by its rows $M_i(j)$:
\begin{equation}
    M_i(j) = \frac{\mathrm{sign}(\mathbf{l}_j^T \Vanishing_i)}{\sqrt{l_j(0)^2 + l_j(1)^2}} \mathbf{l}_j^T,
\end{equation}
One can re-write the previous objective as $M \Vanishing_i = 0$, and the solution is obtained by computing the right null space of the SVD of matrix $M$. This solution becomes our refined vanishing point.

Next, we compute the unknown focal length. From equations (1) and (3) of the main paper and re-using the same notations, we have, for every pair of vanishing points $\Vanishing_i$, $\Vanishing_j$: $\Dir_i^\text{T} \Dir_j = \Vanishing_i^\text{T} (\Intrinsic^{-1})^\text{T} \Intrinsic^{-1} \Vanishing_j = 0$. This constraint can be rewritten as:
\begin{equation}
    -\Vanishing_i(2)\Vanishing_j(2) f^2 = \Vanishing_i(0)\Vanishing_j(0) + \Vanishing_i(1)\Vanishing_j(1),
\end{equation}
where the numbers in the parentheses refer to the $x$, $y$, and $w$ coordinates of the homogeneous points.
Taking $(i,j) \in \{ (1,2), (1,3), (2,3) \}$ gives three independent constraints that all are linear in $f^2$:
\begin{equation}
    \begin{pmatrix}
        -\Vanishing_1(2)\Vanishing_2(2) \\
        -\Vanishing_1(2)\Vanishing_3(2) \\
        -\Vanishing_2(2)\Vanishing_3(2)
    \end{pmatrix} f^2 = \begin{pmatrix}
        \Vanishing_1(0)\Vanishing_2(0) + \Vanishing_1(1)\Vanishing_2(1) \\
        \Vanishing_1(0)\Vanishing_3(0) + \Vanishing_1(1)\Vanishing_3(1) \\
        \Vanishing_2(0)\Vanishing_3(0) + \Vanishing_2(1)\Vanishing_3(1)
    \end{pmatrix}
\end{equation}
We solve it via QR decomposition, and since $f$ can not be negative, we finally obtain $f = +\sqrt{f^2}$.

Finally, we correct the vanishing points to be orthogonal. 
We compute the calibrated vanishing points $\Dir_i, i \in \{1,2,3\}$ as $\Dir_i = \matr K^{-1} \Vanishing_i$, build a matrix $\matr{D} = [\Dir_1 \ \Dir_2 \ \Dir_3]$, and decompose as $\matr{D} = \matr{U}\matr{S}\matr{V}^\text{T}$. We take the rows of matrix $\Rot = \matr{U}\matr{V}^\text{T}$ as the refined calibrated vanishing points.

\section{Alternative Solvers}
\label{sec:derivation_details}

\subsection{Different Elimination Order for \ooz\ Solver}\label{sec:110_alternative}
Here, we give an alternative elimination order for our \ooz\ solver, introduced in Section~\ref{sec:110} of the main paper. We reuse here the same notations. Estimating the orthogonal vanishing points from projections of two mutually orthogonal horizontal lines, and a known vertical direction (referred to as \ooz), leads to two equations of unknown $t$ and $f$:
\begin{equation}
    \begin{split}
        (1-t^2) \Line_1^\text{T} \Intrinsic \Basis_1 - 2t \Line_1^\text{T} \Intrinsic \Basis_2 = 0,\\
        2t \Line_2^\text{T} \Intrinsic \Basis_1 + (1-t^2) \Line_2^\text{T} \Intrinsic \Basis_2 = 0.
    \end{split}
\end{equation}
In the main paper, we propose to solve these equations by first eliminating $t$, and solving for $f$, leading to a degree 2 polynomial. Here, we give an alternative approach, where the equations are solved by eliminating $f$ and solving for $t$ afterwards.

Both equations are linear in $f$. Reusing the notation of the main paper defined in~\eqref{eq:notations}, these two equations become:
\begin{equation}
    \begin{split}
        (1-t^2) (f \delta_1 + \delta_2) - 2 t (f \delta_3 + \delta_4) = 0,\\
        (1-t^2) (f \delta_5 + \delta_6) + 2 t (f \delta_7 + \delta_8) = 0.    
    \end{split}
    \label{eq:f_linear}
\end{equation}
We can easily express the focal length from the first equation of \eqref{eq:f_linear} as a function of $t$ and substitute it into the second equation to get the following constraint:
\begin{equation}
    \begin{split}
        0 = & 4t^2 (\delta_4\delta_7 - \delta_3\delta_8) \\
            & +2t(1 - t^2) (\delta_4\delta_5 + \delta_1\delta_8 - \delta_3\delta_6 - \delta_2\delta_7) \\
            & +(1 - t^2) (\delta_1\delta_6 - \delta_2\delta_5) .
    \end{split}
\end{equation}
This is a univariate quartic polynomial equation. We use the hidden variable approach to solve for $t$, which gives us $4$ solutions. For every solution, we find $f$ from \eqref{eq:f_linear} and only keep $(t,f)$ pairs where the focal length $f$ is positive. Then, we use $t$ to calculate the rotation matrix $\matr{R}$. However, this approach is slower than the one proposed in the main paper: it needs $1.64 \mu s$ to solve one instance of the \ooz\ problem, while the proposed one only needs $0.17 \mu s$.

\subsection{Non Minimal Solver with Linearized Rotation}

In the main paper, we use a non-minimal solver that estimates each vanishing point separately, and then corrects the vanishing points to be orthogonal. Here, we propose an alternative non-minimal solver, which is iterative and uses a first-order approximation of the matrix $\matr{K}\matr{R}$ to estimate all 3 vanishing points simultaneously.

We use the same notation as in the main paper: $\Vanishing_i, i \in \{1,2,3\}$ denotes a vanishing point in direction $\Dir_i$. $\mathcal{L}_i = \{\matr{l}_{i,j}, j \in \{1,...,n_i\} \}, i \in \{1,2,3
\}$ is a set of $n_i$ lines consistent with vanishing point $\mathbf{v}_i$.

This non-minimal solver uses a linearized model of matrix $[\Vanishing_1 \ \Vanishing_2 \ \Vanishing_3] = \Intrinsic \Rot$ to simultaneously  minimize the sum of squared errors:
\begin{equation}
    \sum_{i=1}^3 \sum_{j=1}^{n_i} \left( \frac{\Line_{i,j}^\text{T} \Vanishing_i}{\lVert\Line_{i,j}\rVert} \right)^2. \label{eq:NMS_task}
\end{equation}
Let $\Intrinsic_0 \Rot_0$ be the initial estimate of $\Intrinsic \Rot$, obtained by the minimal solver. The first-order Taylor polynomial of $\Intrinsic \Rot$ can be obtained as:
\begin{equation}
    \Intrinsic \Rot \approx \Intrinsic_0 \Rot_0 + \matr{\delta K} \Rot_0 - \matr{K} \matr{\delta R} \Rot_0
\end{equation}
with derivatives $\matr{\delta K}$ and $\matr{\delta R}$ defined as:
\begin{equation}
    \matr{\delta K} = \begin{bmatrix}
        \delta f & 0 & 0 \\ 0 & \delta f & 0 \\ 0 & 0 & 0
    \end{bmatrix},
    \matr{\delta R} = \begin{bmatrix}
        0 & -\delta v_3 & \delta v_2 \\ \delta v_3 & 0 & -\delta v_1 \\ -\delta v_2 & \delta v_1 & 0
    \end{bmatrix}.
\end{equation}
Now, we can approximate every vanishing point $\Vanishing_i$ as:
\begin{equation}
    \Vanishing_i \approx \matr{B}_i \mathbf{\delta x} + \mathbf{c}_i, \label{eq:NMS_vanishing}
\end{equation}
where
\begin{equation}
\begin{split}
    \matr{B}_i = \begin{bmatrix}
        0 & -f_0 r_{3,i} & f_0 r_{2,i} & r_{1,i}\\
        f_0 r_{3,i} & 0 & -f_0 r_{1,i} & r_{2,i}\\
        -r_{2,i} & r_{1,i} & 0 & 0
    \end{bmatrix},
    \\
    \matr{c}_i = \begin{bmatrix}
        f_0 r_{1,i} \\  f_0 r_{2,i} \\ r_{3,i}
    \end{bmatrix},
    \matr{\delta x} = \begin{bmatrix}
        \delta v_1 & \delta v_2 & \delta v_3 & \delta f
    \end{bmatrix}^\text{T}.
\end{split}    
\end{equation}
We use matrices $\matr{A}_i, i \in \{1,2,3\}$ defined in Section~\ref{sec:NMS} of the main paper, and build matrices:
\begin{equation}
    \matr{A} = \begin{bmatrix}
        \matr{A}_1 & \matr{0} & \matr{0}\\
        \matr{0} & \matr{A}_2 & \matr{0}\\
        \matr{0} & \matr{0} & \matr{A}_2
    \end{bmatrix}, 
    \matr{B} = \begin{bmatrix}
        \matr{B}_1 \\ \matr{B}_2 \\ \matr{B}_3
    \end{bmatrix}, 
    \matr{C} = \begin{bmatrix}
        \matr{c}_1 \\ \matr{c}_2 \\ \matr{c}_3
    \end{bmatrix}.
\end{equation}
Then, we re-write the minimization problem of \eqref{eq:NMS_task} as:
\begin{equation}
    \min \lVert \matr{A}\matr{B}\matr{\delta x} + \matr{A}\matr{C} \rVert^2,
\end{equation}
and we find the update $\matr{\delta x}$ with the least-squares method. Then, we find the vanishing points by \eqref{eq:NMS_vanishing}. To find the focal length $f$ and the orthogonal calibrated vanishing points $\Dir_1, \Dir_2, \Dir_3$ from the uncalibrated vanishing points $\Vanishing_1$, $\Vanishing_2$, $\Vanishing_3$, we use the procedure proposed in the main paper in Section~\ref{sec:NMS}.
For the next iteration, we set $f_0 := f$, and $\Rot_0 := [\Dir_1, \Dir_2, \Dir_3]$, and find the update in the same way.

The comparison of this non-minimal solver with the one proposed in the main paper is shown in the following section. We observed a lower performance for the solver with linearized rotation in our real-world experiments, and thus only presented the non orthogonal one in the main paper. However, the former could still be used in cases with small rotations with only a few iterations, and could become an efficient alternative to the non orthogonal solver.

\section{Additional Synthetic Tests}
\label{sec:synthetic_tests}

In order to further evaluate the solvers in various scenarios, we have performed additional synthetic tests, presented in this section.

\begin{figure}
    \centering
    \setlength\tabcolsep{0pt}
    \setlength\extrarowheight{-3pt}
    \includegraphics[width=0.9\linewidth]{figs/legend_solvers.png}
    \begin{tabular}{c c}
       \includegraphics[width=0.5\linewidth]{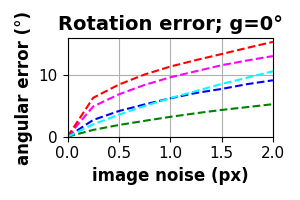}
       &
       \includegraphics[width=0.5\linewidth]{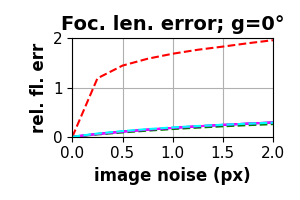}
       \\

       \includegraphics[width=0.5\linewidth]{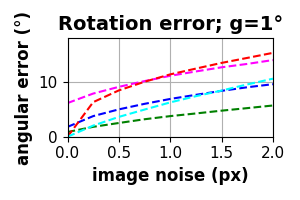}
       &
       \includegraphics[width=0.5\linewidth]{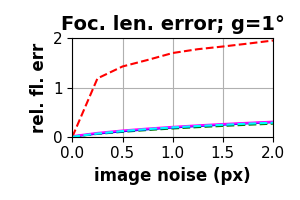}
       \\

       \includegraphics[width=0.5\linewidth]{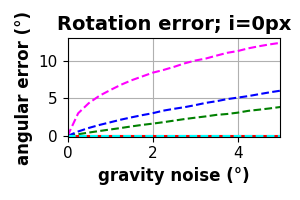}
       &
       \includegraphics[width=0.5\linewidth]{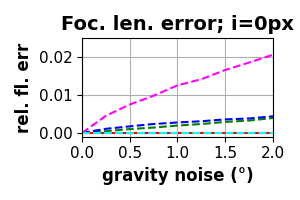}
       \\

        \includegraphics[width=0.5\linewidth]{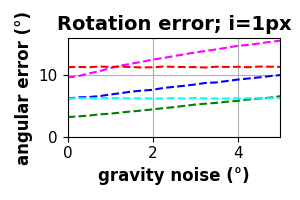}
       &
       \includegraphics[width=0.5\linewidth]{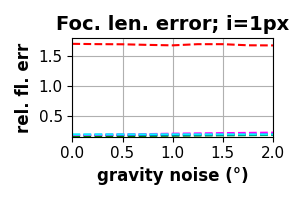}
       \\

    \end{tabular}
    
    \caption{\textbf{Noise study of the linearized non-minimal solver.}: The average (over 100000 runs) rotation (left), and relative focal length error (right) of five solvers as a function of the image and gravity noise. The results are refined with a linearized non-minimal solver with $10$ iterations and $20$ lines per direction. The fixed noise std. is in the title (g - gravity, i - image).}
    \label{fig:nms_tests_all}
\end{figure}

\begin{figure}
    \centering
    \setlength\tabcolsep{0pt}
    \setlength\extrarowheight{-3pt}
    \includegraphics[width=0.9\linewidth]{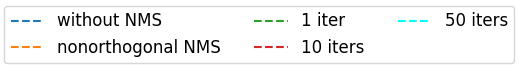}
    \begin{tabular}{c c}
       \includegraphics[width=0.5\linewidth]{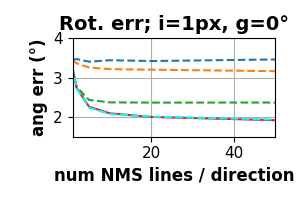}
       &
       \includegraphics[width=0.5\linewidth]{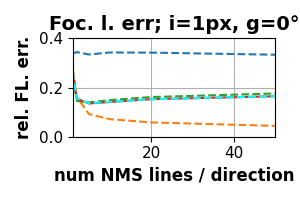}
       \\

       \includegraphics[width=0.5\linewidth]{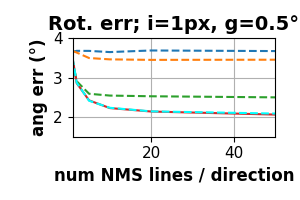}
       &
       \includegraphics[width=0.5\linewidth]{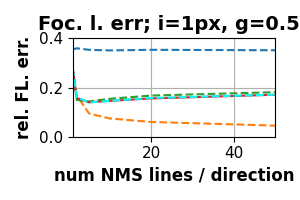}
       \\

       \includegraphics[width=0.5\linewidth]{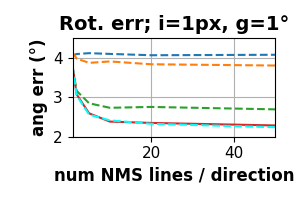}
       &
       \includegraphics[width=0.5\linewidth]{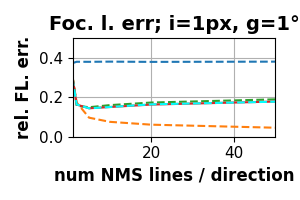}
       \\
    \end{tabular}
    
    \caption{\textbf{Effect of the number of lines on non-minimal solvers (NMS).}: Average (over $100000$ samples) rotation (left) and relative focal length error (right) as a function of the number of lines used in the NMS. We considered the following approaches: baseline (without NMS) , non-orthogonal NMS presented in the main paper, and linearized rotation with $n$ iterations ($n$ iters). The \ooz\ solver was used for the initialization. The fixed noise std. is in the title (g - gravity, i - image).}
    \label{fig:nms_tests_lines_iters}
\end{figure}

\begin{figure}
    \centering
    \setlength\tabcolsep{0pt}
    \setlength\extrarowheight{-3pt}
    \includegraphics[width=0.9\linewidth]{figs/legend_grid.png}
    \begin{tabular}{c c}
       \includegraphics[width=0.5\linewidth]{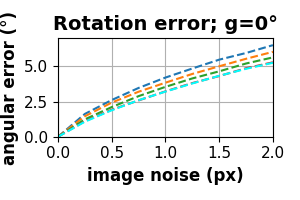}
       &
       \includegraphics[width=0.5\linewidth]{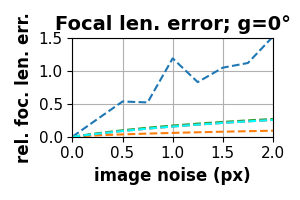}
       \\

       \includegraphics[width=0.5\linewidth]{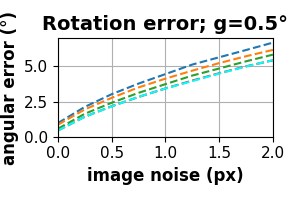}
       &
       \includegraphics[width=0.5\linewidth]{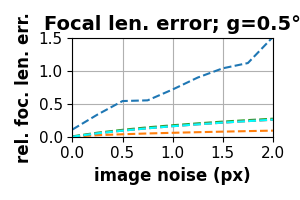}
       \\

       \includegraphics[width=0.5\linewidth]{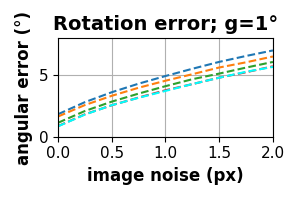}
       &
       \includegraphics[width=0.5\linewidth]{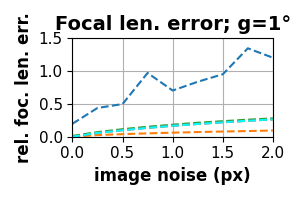}
       \\

       \includegraphics[width=0.5\linewidth]{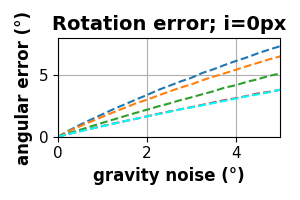}
       &
       \includegraphics[width=0.5\linewidth]{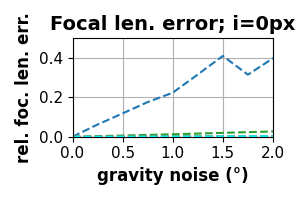}
       \\

        \includegraphics[width=0.5\linewidth]{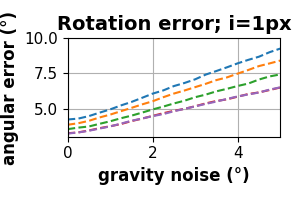}
       &
       \includegraphics[width=0.5\linewidth]{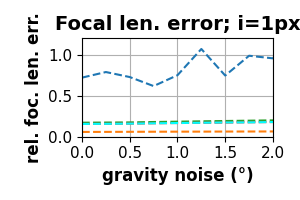}
       \\

        \includegraphics[width=0.5\linewidth]{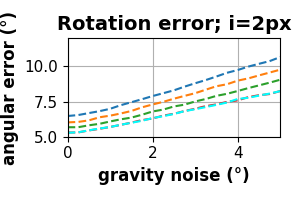}
       &
       \includegraphics[width=0.5\linewidth]{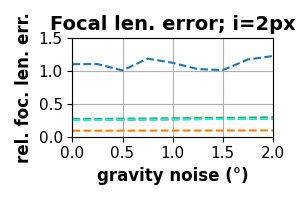}
       \\

    \end{tabular}
    
    \caption{\textbf{Effect of noise on our non-minimal solvers (NMS).}: The plots show the relation between the noise of the input (endpoints of the lines in px, gravity vector in degrees) and the rotation and relative focal length error. We consider the following approaches: baseline (without NMS) , non orthogonal NMS of the mai paper, linearized rotation with $n$ iterations ($n$ iters). The \ooz\ solver was used for the initialization. The fixed noise std. is in the title (g - gravity, i - image).}
    \label{fig:nms_tests_first_order}
\end{figure}

\begin{figure}
    \centering
    \includegraphics[width=0.9\linewidth]{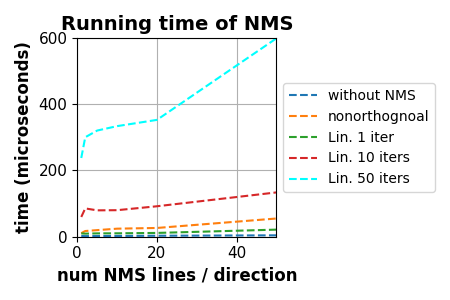}
    \caption{\textbf{Run-time of our non-minimal solvers} as a function of number of lines used in the solver.}
    \label{fig:running_time}
\end{figure}

\textbf{Non minimal solver tests.} To evaluate the non-minimal solvers, we generated the minimal problems exactly as in the main paper. Figure \ref{fig:nms_tests_all} shows the average rotation and focal length errors of every proposed minimal solvers refined by the linearized non-minimal solver with $10$ iterations. The \ooz\ solver leads to the most accurate solutions on almost all noise configurations. Figure \ref{fig:nms_tests_lines_iters} shows the average rotation and focal length errors of the results of different non-minimal solvers, as a function of the number of lines used within the solver. The error does not change significantly after adding more than $20$ lines per direction. While the linearized non-minimal solver gives a better estimation of the rotation, using the nonorthogonal non-minimal solver leads to lower focal length errors. Figure \ref{fig:nms_tests_first_order} shows the errors of different non-minimal solvers as a function of the input noise. Again, the linearized non-minimal solver gives more accurate rotations, while the nonorthogonal non-minimal solver gives more accurate focal lengths. Figure \ref{fig:running_time} shows the evaluation of the running time of different non-minimal solvers. The runtime increases with the increasing number of lines used within the non-minimal solver, and this growth is roughly linear. 

\begin{figure}
    \centering
    \setlength\tabcolsep{0pt}
    \setlength\extrarowheight{-3pt}
    \includegraphics[width=0.9\linewidth]{figs/legend_solvers.png}
    \begin{tabular}{c c}
       \includegraphics[width=0.5\linewidth]{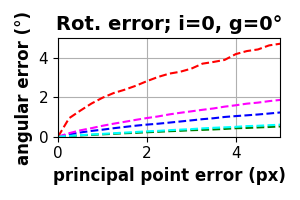}
       &
       \includegraphics[width=0.5\linewidth]{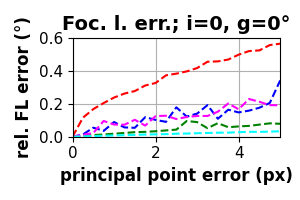}
       \\
       \includegraphics[width=0.5\linewidth]{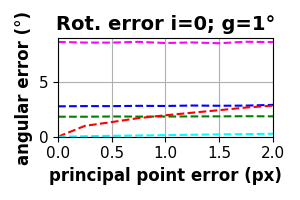}
       &
       \includegraphics[width=0.5\linewidth]{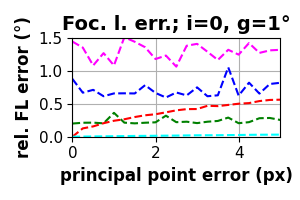}
       \\
       \includegraphics[width=0.5\linewidth]{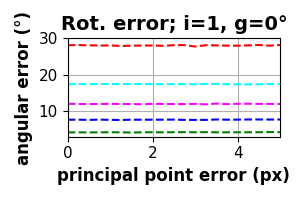}
       &
       \includegraphics[width=0.5\linewidth]{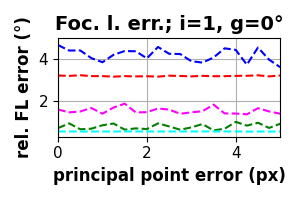}
       \\
    \end{tabular}
    
    \caption{\textbf{Effect of noise on the principal point.} The average (over 100000 runs) rotation (left), and relative focal length error (right) of the proposed solvers as a function of the principal point noise $\sigma_p$. The fixed noise std. is in the title (g - gravity, i - image).}
    \label{fig:pp_noise_tests}
\end{figure}

\begin{figure}
    \centering
    \includegraphics[width=0.9\linewidth]{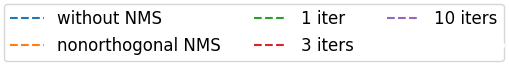}
    \begin{tabular}{c c}
       \includegraphics[width=0.5\linewidth]{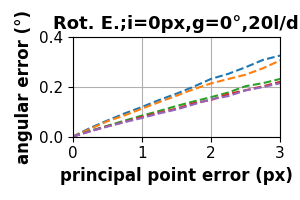}
       &
       \includegraphics[width=0.5\linewidth]{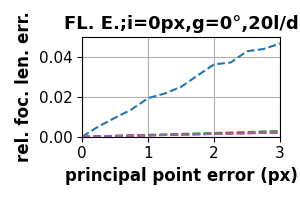}
       \\
       \includegraphics[width=0.5\linewidth]{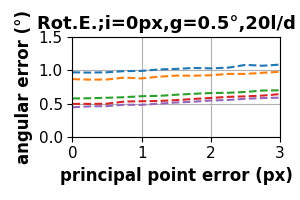}
       &
       \includegraphics[width=0.5\linewidth]{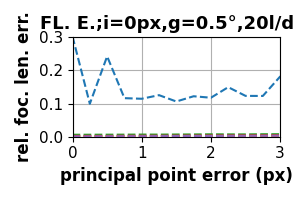}
       \\
       \includegraphics[width=0.5\linewidth]{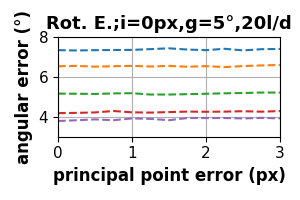}
       &
       \includegraphics[width=0.5\linewidth]{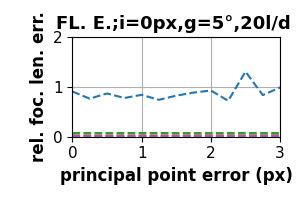}
       \\
       
       \includegraphics[width=0.5\linewidth]{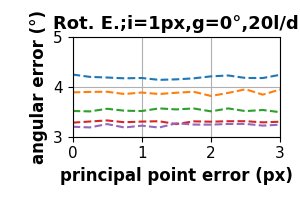}
       &
       \includegraphics[width=0.5\linewidth]{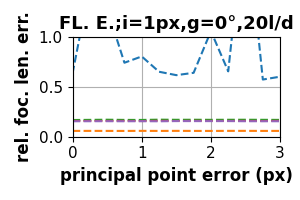}
       \\
       \includegraphics[width=0.5\linewidth]{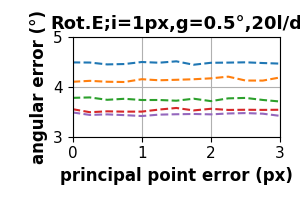}
       &
       \includegraphics[width=0.5\linewidth]{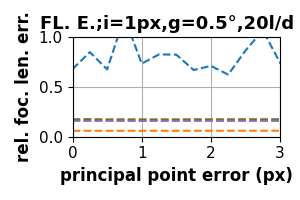}
       \\
       \includegraphics[width=0.5\linewidth]{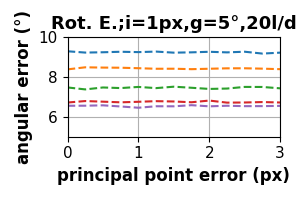}
       &
       \includegraphics[width=0.5\linewidth]{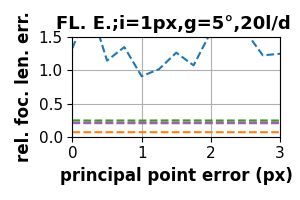}
       \\

    \end{tabular}
    
    \caption{\textbf{Relation between the principal point noise and our non-minimal solvers. } The plots show the relation between the error of the principal point and the rotation and relative focal length error. We considered the following approaches: baseline (without NMS) , non orthogonal NMS as in the main paper, and linearized rotation with $n$ iterations ($n$ iters). The \ooz\ solver was used for the initialization. The fixed noise std. is in the title (g - gravity, i - image).}
    \label{fig:PP_nms_tests_first_order}
\end{figure}

\textbf{Principal point tests.} To evaluate the robustness of our solvers to the incorrectly estimated principal point, we generate minimal problems similarly to the main paper, and we perturb the principal point with Gaussian noise with standard deviation $\sigma_p$. Figure \ref{fig:pp_noise_tests} shows the average rotation and focal length errors on different levels of $\sigma_p$. Solvers \ooz\ and 2-2-0 lead to the more accurate solutions. Figure \ref{fig:PP_nms_tests_first_order} shows the average rotation and focal length errors on different levels of $\sigma_p$ refined by the non-minimal solver. It can be seen that all the proposed non-minimal solvers are very robust to small noise on the principal point, and that their estimate of the focal length is significantly better than without any local optimization.

\section{Additional Ablations on the LO}
\label{sec:lo_supp}

\noindent \textbf{Evaluation on ScanNet~\cite{dai2017scannet}.}
Table~\ref{tab:scannet_lo} shows a similar ablation study for our proposed Local Optimization (LO) as in the main paper, but for the ScanNet dataset~\cite{dai2017scannet}.
Note that the lack of improvement from Iter for our solvers stems from the very noisy gravity prior in this dataset, that makes the initial model too noisy for the optimization to converge to an accurate solution.

\begin{table}[]
    \centering
    \scriptsize
    \setlength{\tabcolsep}{5pt}
    \begin{tabular}{ccccccc}
        \toprule
        \multirow{2}{*}{Solver} & \multirow{2}{*}{LO} & \multicolumn{2}{c}{Rotation estimation} & \multicolumn{2}{c}{VP estimation} & \multirow{2}{*}{\makecell{Time\\(ms)}} \\
        \cmidrule(lr){3-4} \cmidrule(lr){5-6}
         & & Err $\downarrow$ & AUC $\uparrow$ & Err $\downarrow$ & AUC $\uparrow$ &  \\
        \midrule
        \multirow{3}{*}{\makecell{2-2-0~\cite{wildenauer2012}\\No gravity}} & No & 42.98 & \phantom{1}0.7 / \phantom{1}3.6 / 11.6 & 8.45 & 1.94 & \textbf{22} \\
        & Iter~\cite{zhang2015} & \textbf{42.96} & \phantom{1}3.5 / \phantom{1}8.7 / 16.5 & 7.55 & 2.73 & 26 \\
         & Ours & 43.47 & \textbf{\phantom{1}6.8 / 13.8 / 21.5} & \textbf{6.50} & \textbf{3.67} & 130 \\
        \cmidrule(lr){1-2}
        \multirow{3}{*}{\makecell{2-1-1~\cite{wildenauer2012}\\No gravity}} & No & 42.21 & \phantom{1}1.0 / \phantom{1}4.9 / 14.3 & 8.09 & 2.26 & \textbf{28} \\
        & Iter~\cite{zhang2015} & 42.69 & \phantom{1}4.0 / 10.2 / 18.1 & 7.18 & 2.99 & 33 \\
         & Ours & \textbf{42.07} & \textbf{\phantom{1}7.0 / 14.3 / 22.0} & \textbf{6.42} & \textbf{3.75} & 112 \\
        \cmidrule(lr){1-2}
        \multirow{3}{*}{\makecell{\tzz~\cite{lobo2003}\\+ Prior}} & No & 50.07 & \phantom{1}0.2 / \phantom{1}0.7 / \phantom{1}2.8 & 9.40 & 0.64 & \textbf{24} \\
        & Iter~\cite{zhang2015} & 50.46 & \phantom{1}0.2 / \phantom{1}0.7 / \phantom{1}2.9 & 9.39 & 0.64 & 30 \\
         & Ours & \textbf{43.99} & \textbf{\phantom{1}6.5 / 13.1 / 20.2} & \textbf{6.60} & \textbf{3.55} & 131 \\
        \cmidrule(lr){1-2}
        \multirow{3}{*}{\makecell{\zoo\\+ Prior}} & No & 49.87 & \phantom{1}0.2 / \phantom{1}0.7 / \phantom{1}2.8 & 9.40 & 0.64 & \textbf{14} \\
        & Iter~\cite{zhang2015} & 50.01 & \phantom{1}0.3 / \phantom{1}1.0 / \phantom{1}3.2 & 9.32 & 0.71 & 18 \\
         & Ours & \textbf{42.59} & \textbf{\phantom{1}5.8 / 12.2 / 19.9} & \textbf{6.74} & \textbf{3.49} & 247 \\
        \cmidrule(lr){1-2}
        \multirow{3}{*}{\makecell{\ooz\\+ Prior}} & No & 50.49 & \phantom{1}0.6 / \phantom{1}1.7 / \phantom{1}4.4 & 9.30 & 0.73 & \textbf{9} \\
        & Iter~\cite{zhang2015} & 50.39 & \phantom{1}1.2 / \phantom{1}2.9 / \phantom{1}6.7 & 8.99 & 1.12 & 16 \\
         & Ours & \textbf{43.22} & \textbf{\phantom{1}6.4 / 12.9 / 20.8} & \textbf{6.64} & \textbf{3.49} & 273 \\
        \cmidrule(lr){1-2}
        \multirow{3}{*}{\makecell{Hybrid\\+ Prior}} & No & 43.19 & \phantom{1}0.8 / \phantom{1}2.8 / \phantom{1}9.2 & 8.66 & 1.80 & \textbf{8} \\
        & Iter~\cite{zhang2015} & 42.85 & \phantom{1}2.6 / \phantom{1}7.3 / 15.2 & 7.64 & 2.73 & 17 \\
         & Ours & \textbf{42.72} & \textbf{\phantom{1}7.9 / 16.2 / 24.8} & \textbf{5.89} & \textbf{4.21} & 160 \\
        \bottomrule
    \end{tabular}
    \caption{\textbf{Local optimization on the ScanNet dataset~\cite{dai2017scannet}.} We report median errors in degrees and AUCs over 10 runs. The rotation AUC is given at thresholds 5$^\circ$ / 10$^\circ$ / 20$^\circ$.}
    \label{tab:scannet_lo}
\end{table}

\noindent \textbf{Number of Iterations.}
The results in the main paper correspond to the best numbers that could be obtained, assuming that time is not a limitation. In resource-constrained scenarios when speed matters, it is also possible to reduce the number of LO iterations to significantly speed up the vanishing point detection, for a minor drop of performance. As shown in Table~\ref{tab:yud_lo_iter}, running only 10 iterations of LO is already enough to achieve a high performance, at a negligible overhead time.

\begin{table}[]
    \centering
    \scriptsize
    \setlength{\tabcolsep}{5pt}
    \begin{tabular}{cccccc}
        \toprule
        \multirow{2}{*}{\makecell{\# LO\\iterations}} & \multicolumn{2}{c}{Rotation estimation} & \multicolumn{2}{c}{VP estimation} & \multirow{2}{*}{\makecell{Time\\(ms)}} \\
        \cmidrule(lr){2-3} \cmidrule(lr){4-5}
        & Err $\downarrow$ & AUC $\uparrow$ & Err $\downarrow$ & AUC $\uparrow$ &  \\
        \midrule
        0 & 4.59 & 26.4 / 46.1 / 64.9 & 4.30 & 5.95 & \textbf{43} \\
        10 & 1.16 & 69.3 / 83.7 / 90.9 & 1.31 & 8.75 & 46 \\
        20 & 1.18 & 70.0 / 84.0 / 91.0 & 1.29 & 8.79 & 55 \\
        100 & \textbf{1.13} & \textbf{71.5 / 85.8 / 92.9} & \textbf{1.16} & \textbf{8.97} & 100 \\
        \bottomrule
    \end{tabular}
    \caption{\textbf{Study on the number of LO iterations.} We report median errors in degrees and AUCs over 30 runs on the YorkUrban dataset~\cite{yorkurban} for the hybrid solver with prior gravity. The rotation AUC is given 5$^\circ$ / 10$^\circ$ / 20$^\circ$.}
    \label{tab:yud_lo_iter}
\end{table}

\section{Generalization to other RANSAC}
\label{sec:magsac}

While our proposed approach leverages existing RANSAC frameworks such as LO-RANSAC~\cite{Lebeda2012loransac} and hybrid RANSAC~\cite{Camposeco2018CPVR}, it can also be applied to more recent RANSAC strategies. We adapt here our approach to MAGSAC~\cite{barath2019magsac,barath2019magsacplusplus}, one of the state-of-the-art RANSAC currently existing. We replace our scoring method with the one proposed in MAGSAC, and obtain the results of Table~\ref{tab:magsac_ablation}. MAGSAC scoring can slightly boost the performance, showing that future improvements on RANSAC can further benefit our approach.

\begin{table}[h]
    \centering
    \scriptsize
    \setlength{\tabcolsep}{3pt}
    \begin{tabular}{cccccccc}
        \toprule
         & \multirow{2}{*}{\makecell{RANSAC\\scoring}} & \multicolumn{2}{c}{Rotation estimation} & \multicolumn{2}{c}{VP estimation} & \multirow{2}{*}{\makecell{Time\\(ms)}} \\
        \cmidrule(lr){3-4} \cmidrule(lr){5-6}
        &  & Err $\downarrow$ & AUC $\uparrow$ & Err $\downarrow$ & AUC $\uparrow$ &  \\
        \midrule
        \multirow{2}{*}{YorkUrban~\cite{yorkurban}} & RANSAC & 1.12 & 72.5 / 86.3 / 93.2 & 1.11 & \textbf{9.08} & \textbf{79} \\
          & MAGSAC & \textbf{1.11} & \textbf{72.9 / 86.5 / 93.3} & \textbf{1.10} & 9.06 & 81 \\
        \midrule
        \multirow{2}{*}{ScanNet~\cite{dai2017scannet}} & RANSAC & 25.83 & 20.3 / 28.9 / \textbf{36.6} & 3.09 & 6.94 & 190 \\
          & MAGSAC & \textbf{25.75} & \textbf{20.6} / \textbf{29.0} / 36.5 & \textbf{3.02} & \textbf{7.00} & \textbf{178} \\
        \bottomrule
    \end{tabular}
    \caption{\textbf{RANSAC vs MAGSAC.} We report the rotation estimation AUC at $5^\circ / 10^\circ / 20^\circ$ and VP estimation metrics on the YorkUrban~\cite{yorkurban} and ScanNet~\cite{dai2017scannet} datasets for the hybrid solver with IMU gravity.}
    \label{tab:magsac_ablation}
\end{table}

\section{Visualizations of VPs and their Applications}
\label{sec:visualizations}

\begin{figure*}
    \centering
    \small
    \newcommand{\ratio}{0.25}
    \setlength{\tabcolsep}{9pt}
    \begin{tabular}{ccc}
        \textbf{2-1-1 solver~\cite{wildenauer2012}} & \textbf{Hybrid - Prior gravity} & \textbf{Hybrid - Ground truth gravity} \vspace{0.1cm} \\
        \includegraphics[width=\ratio\textwidth]{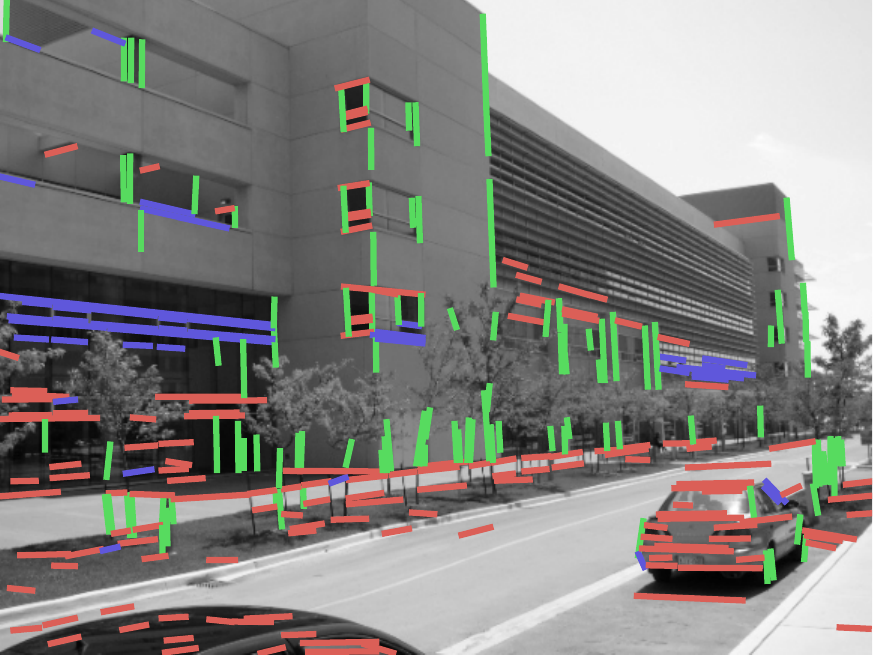}
        & \includegraphics[width=\ratio\textwidth]{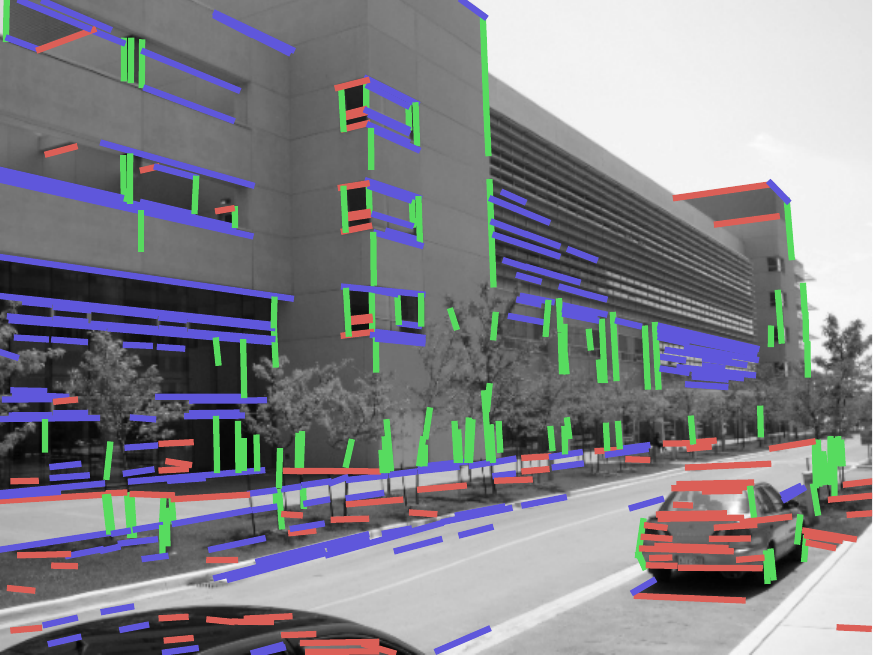}
        & \includegraphics[width=\ratio\textwidth]{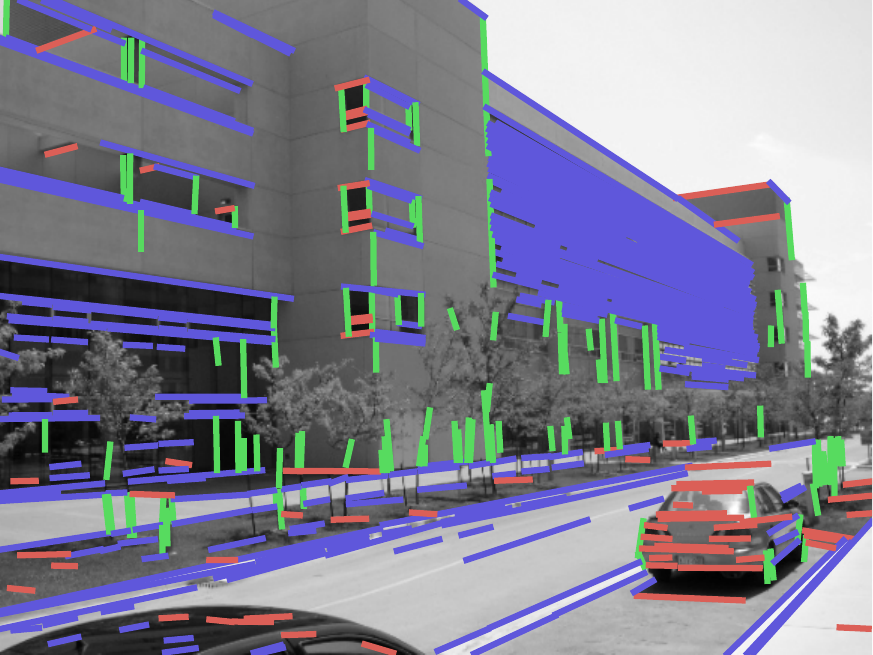} \\
        Num inliers = 292 ; Rot error = 8.92 & Num inliers = 353 ; Rot error = 5.38 & Num inliers = 402 ; Rot error = 0.48 
        \vspace{0.1cm}\\
        \includegraphics[width=\ratio\textwidth]{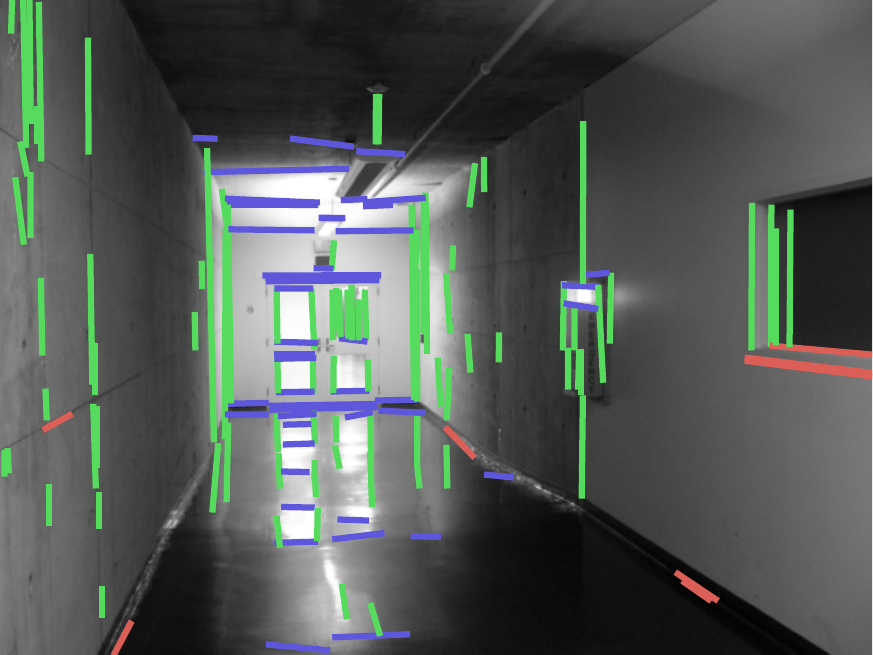}
        & \includegraphics[width=\ratio\textwidth]{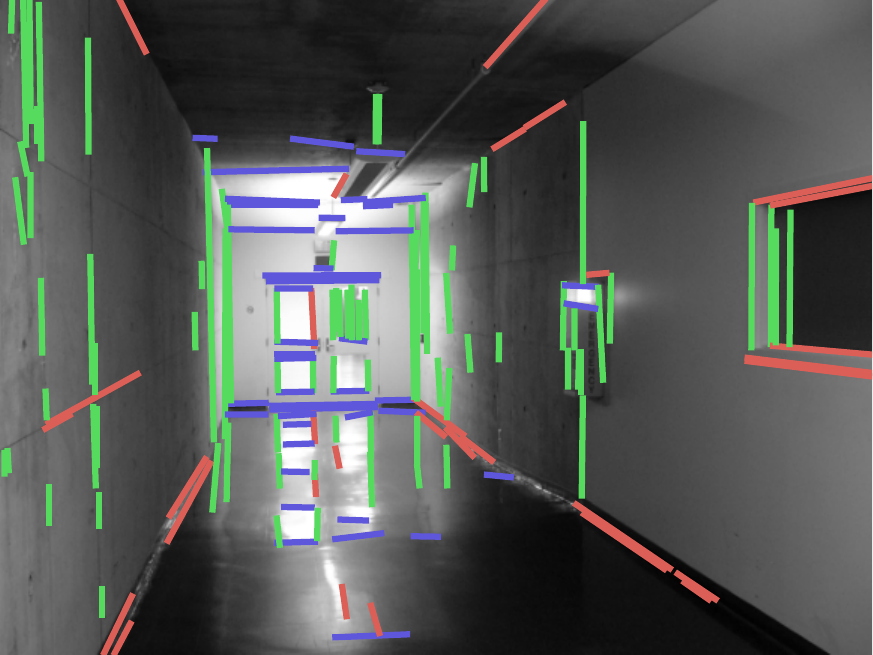}
        & \includegraphics[width=\ratio\textwidth]{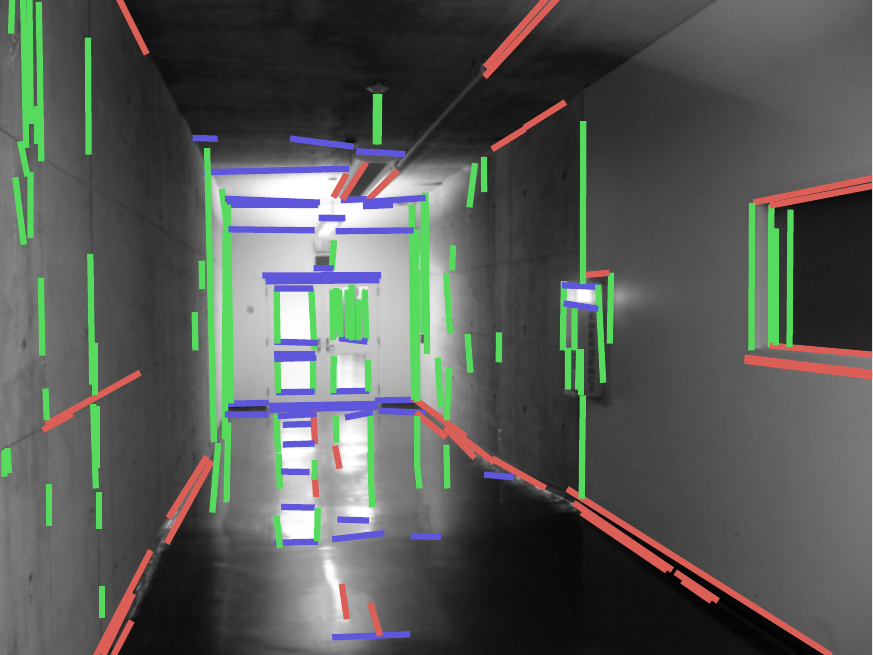} \\
        Num inliers = 138 ; Rot error = 6.81 & Num inliers = 154 ; Rot error = 2.05 & Num inliers = 160 ; Rot error = 1.43 \vspace{0.1cm} \\
        \includegraphics[width=\ratio\textwidth]{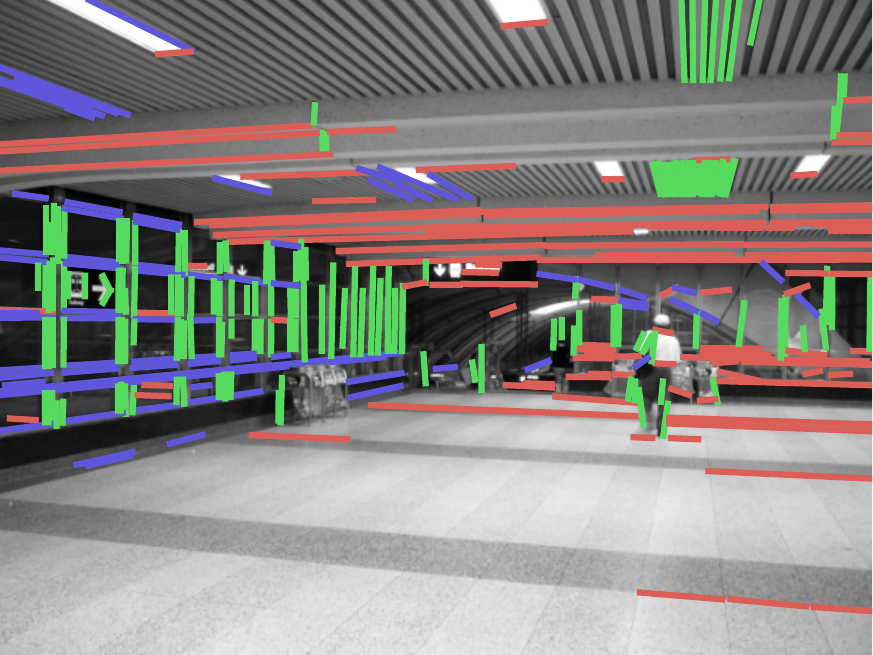}
        & \includegraphics[width=\ratio\textwidth]{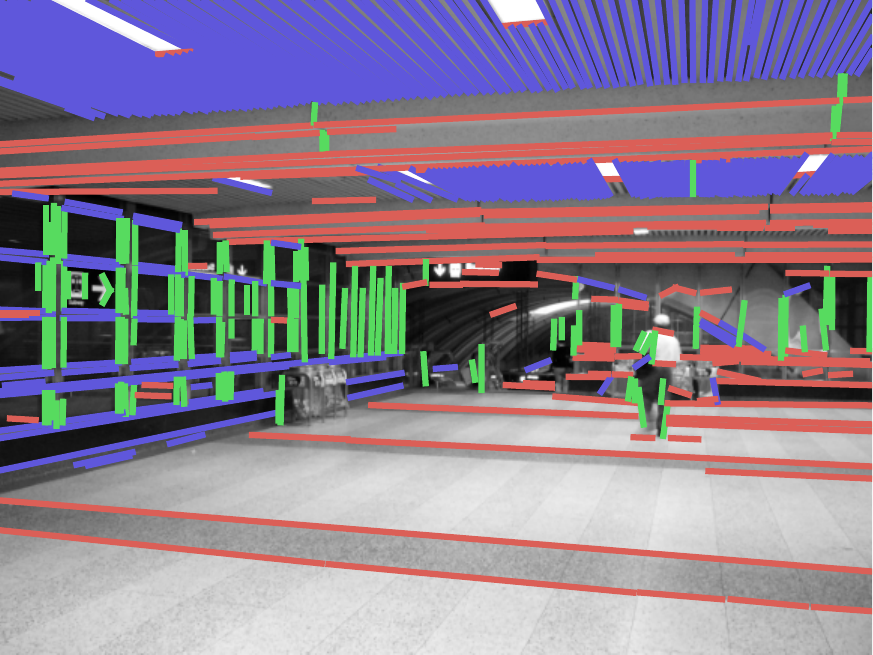}
        & \includegraphics[width=\ratio\textwidth]{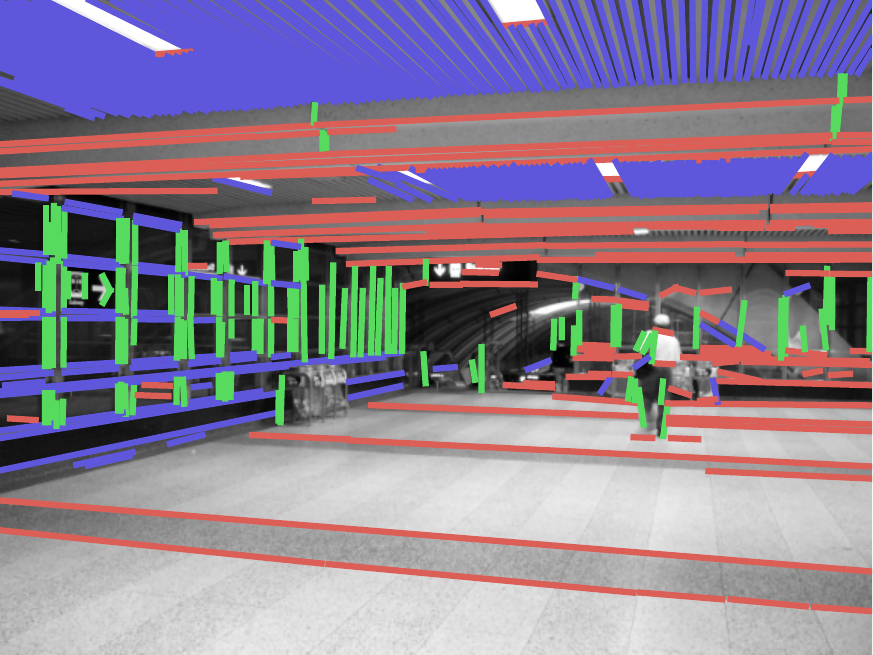} \\
        Num inliers = 324 ; Rot error = 8.55 & Num inliers = 488 ; Rot error = 0.48 & Num inliers = 487 ; Rot error = 0.37 \vspace{0.1cm} \\
        \includegraphics[width=\ratio\textwidth]{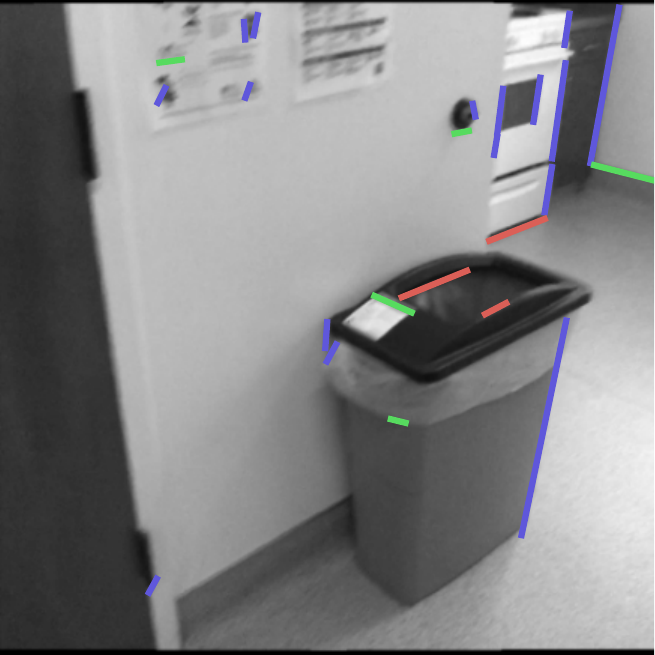}
        & \includegraphics[width=\ratio\textwidth]{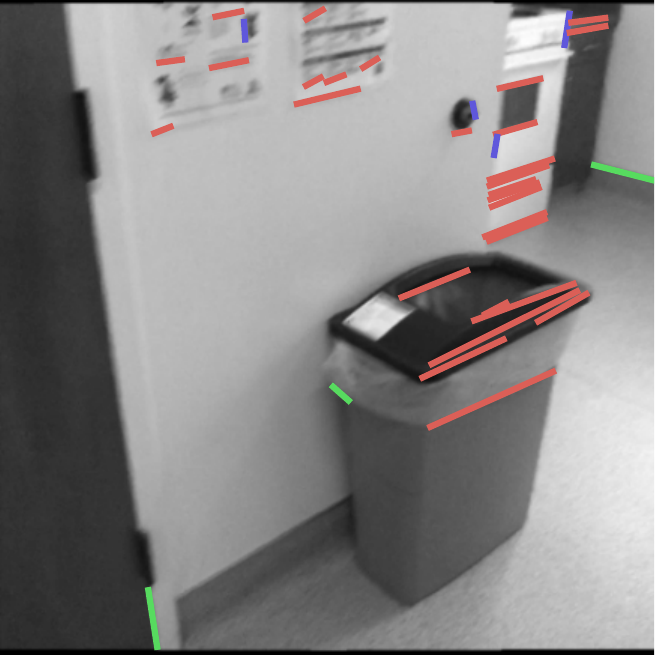}
        & \includegraphics[width=\ratio\textwidth]{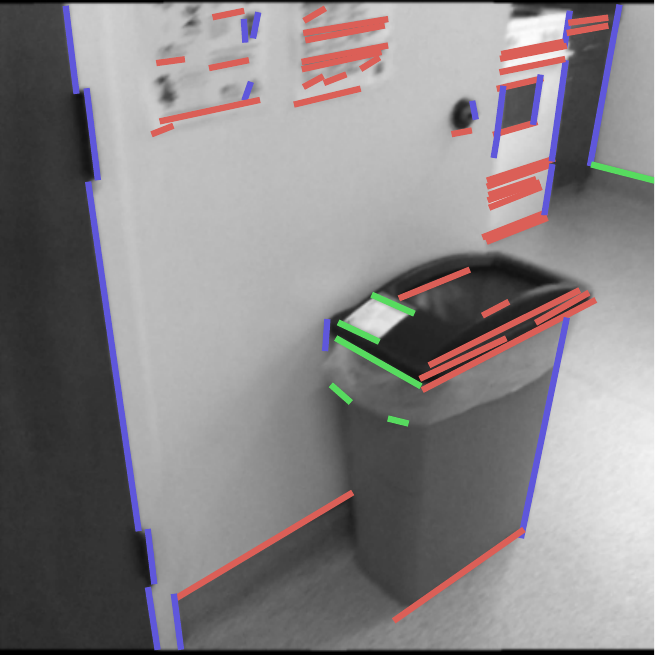} \\
        Num inliers = 24 ; Rot error = 32.68 & Num inliers = 35 ; Rot error = 22.03 & Num inliers = 62 ; Rot error = 2.67 \vspace{0.1cm} \\
        \includegraphics[width=\ratio\textwidth]{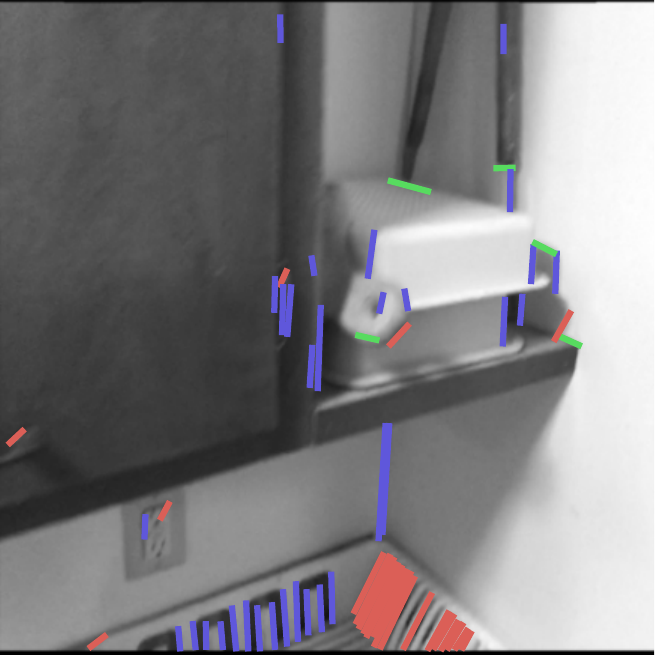}
        & \includegraphics[width=\ratio\textwidth]{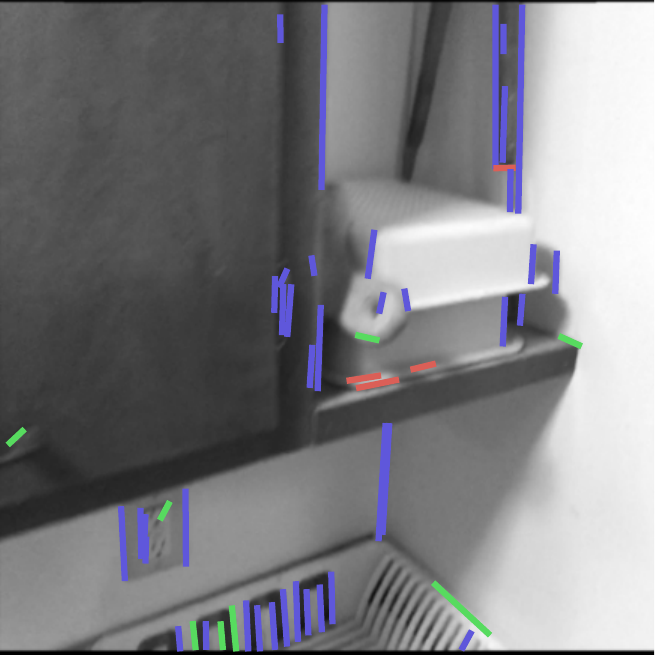}
        & \includegraphics[width=\ratio\textwidth]{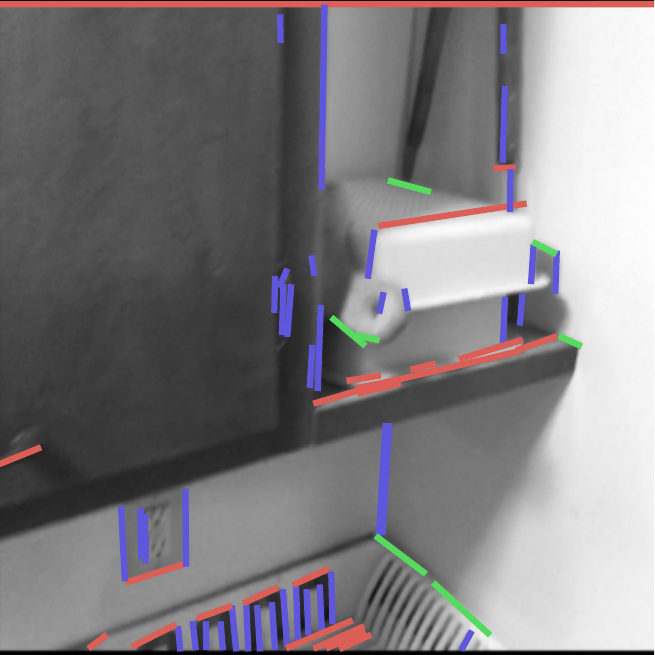} \\
        Num inliers = 58 ; Rot error = 40.45 & Num inliers = 51 ; Rot error = 33.76 & Num inliers = 69 ; Rot error = 1.29 \\
    \end{tabular}
    \caption{\textbf{Visualization of vanishing points.} We display the inlier lines with one color per vanishing point, for the 2-1-1 solver of~\cite{wildenauer2012}, and our hybrid solver with either prior or ground truth gravity. The first three rows are from YorkUrban~\cite{yorkurban}, and last two rows from ScanNet~\cite{dai2017scannet}.}
    \label{fig:visualizations}
\end{figure*}

We display several visualizations of the inlier lines for each vanishing point in Figure~\ref{fig:visualizations}, with one color per VP. Our hybrid RANSAC with prior gravity is able to find more inliers and better rotations than the previous best solver for uncalibrated images~\cite{wildenauer2012}. When using the ground truth gravity, the results are even further improved.

Note that in the last two rows, we display images of the ScanNet dataset~\cite{dai2017scannet}, which has some unstructured scenes and can be quite challenging for vanishing point estimation. For example, in the last row, the 2-1-1 solver~\cite{wildenauer2012} is misleaded by the many red lines of the dish dryer, which are actually inconsistent with the real Manhattan directions. On the contrary, our approach with prior gravity finds a better vertical direction and consequently better recognizes the other two vanishing directions. Finally, the hybrid RANSAC leveraging ground truth gravity is able to detect correctly all the Manhattan directions and obtains a much lower rotation error.

\begin{figure*}
    \centering
    \small
    \newcommand{\ratio}{0.47}
    \newcommand{\ratiolamar}{0.22}
    \setlength{\tabcolsep}{7pt}
    \begin{tabular}{cccc}
        \multicolumn{2}{c}{\textbf{2-1-1 solver~\cite{wildenauer2012}}} & \multicolumn{2}{c}{\textbf{Hybrid solver}} \vspace{0.1cm} \\
        \multicolumn{2}{c}{\includegraphics[width=\ratio\textwidth]{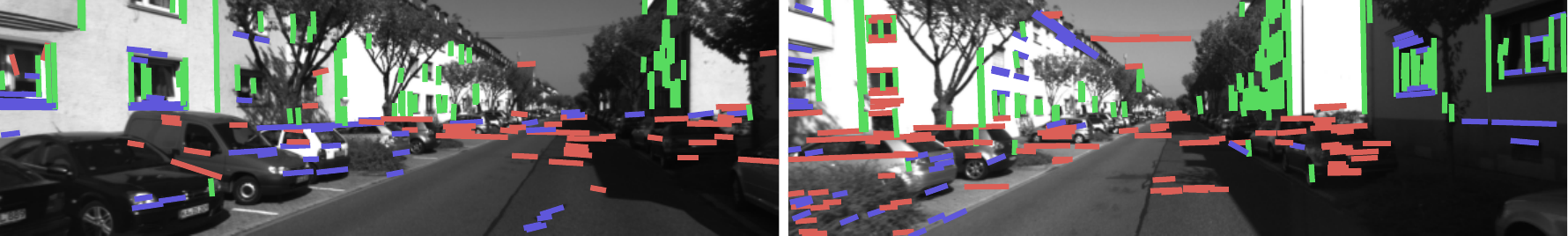}}
        & \multicolumn{2}{c}{\includegraphics[width=\ratio\textwidth]{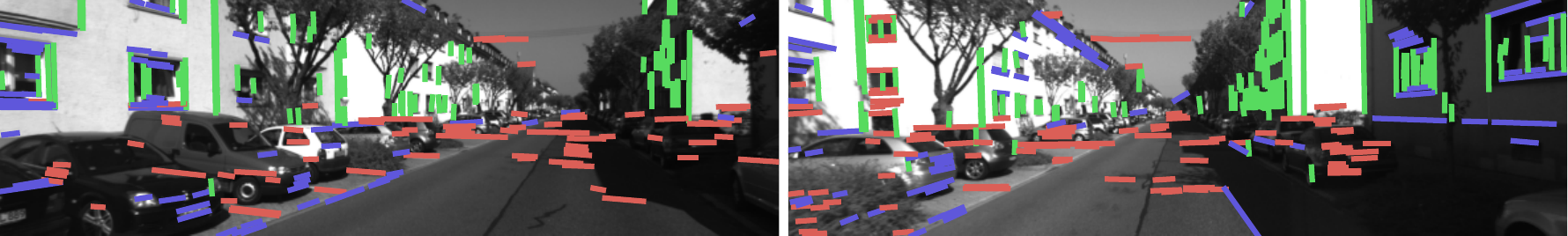}} \\
        \multicolumn{2}{c}{Rot error = 30.58$^\circ$ ;  f error = 0.514} & \multicolumn{2}{c}{Rot error = 6.88$^\circ$ ; f error = 0.290} \vspace{0.1cm}\\
        \multicolumn{2}{c}{\includegraphics[width=\ratio\textwidth]{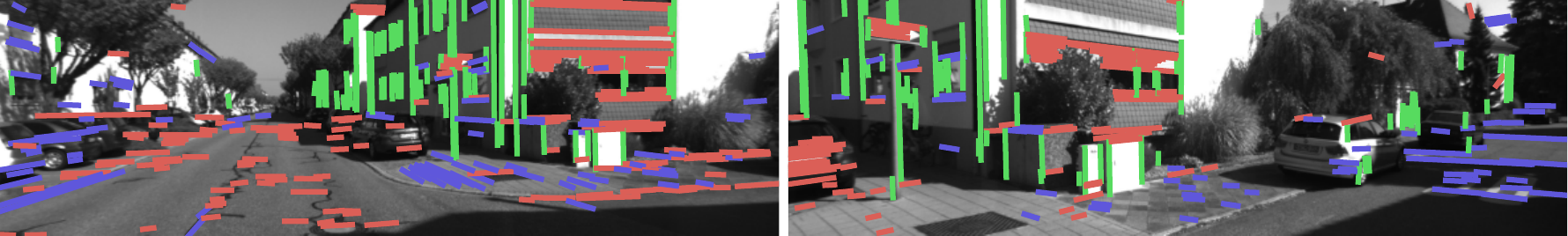}}
        & \multicolumn{2}{c}{\includegraphics[width=\ratio\textwidth]{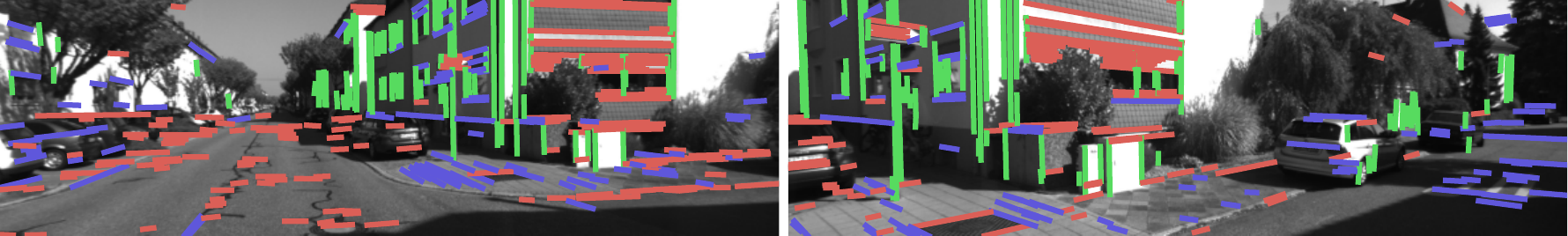}} \\
        \multicolumn{2}{c}{Rot error = 12.13$^\circ$ ;  f error = 0.309} & \multicolumn{2}{c}{Rot error = 4.05$^\circ$ ; f error = 0.182} \vspace{0.1cm}\\
        \multicolumn{2}{c}{\includegraphics[width=\ratio\textwidth]{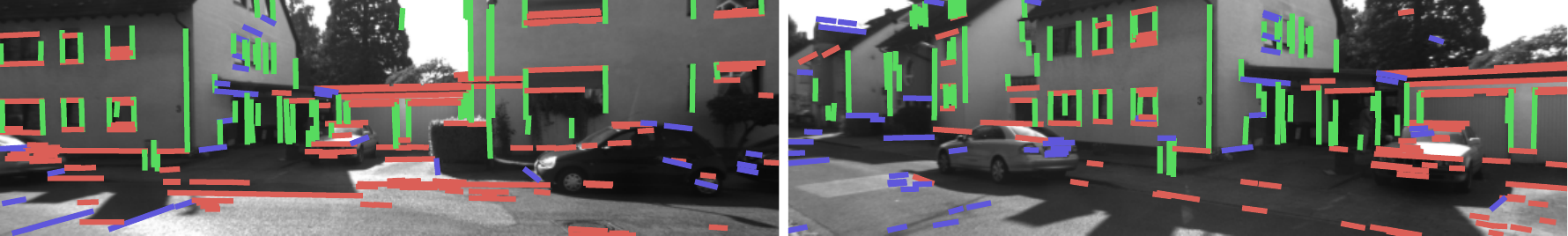}}
        & \multicolumn{2}{c}{\includegraphics[width=\ratio\textwidth]{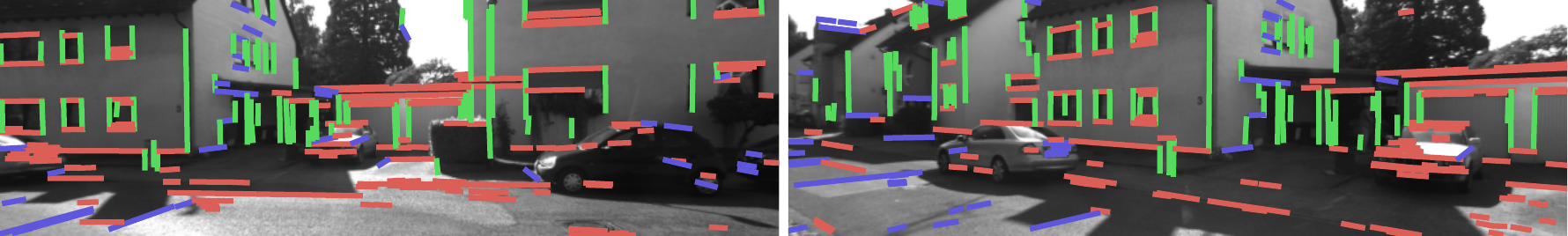}} \\
        \multicolumn{2}{c}{Rot error = 9.59$^\circ$ ;  f error = 0.436} & \multicolumn{2}{c}{Rot error = 2.74$^\circ$ ; f error = 0.071} \vspace{0.5cm}\\
        \includegraphics[width=\ratiolamar\textwidth]{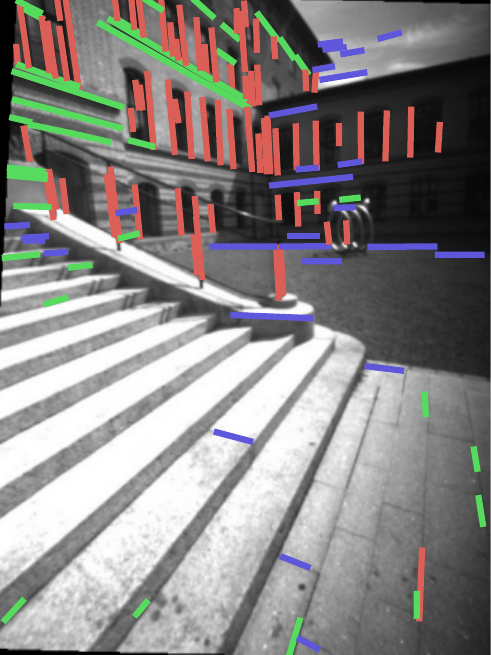}
        & \includegraphics[width=\ratiolamar\textwidth]{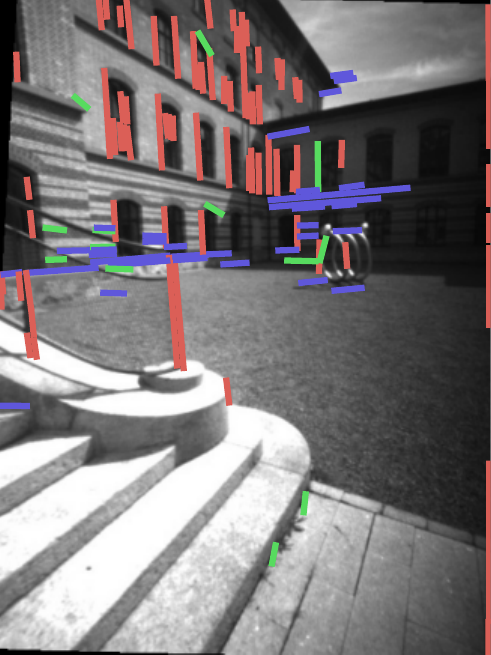}
        & \includegraphics[width=\ratiolamar\textwidth]{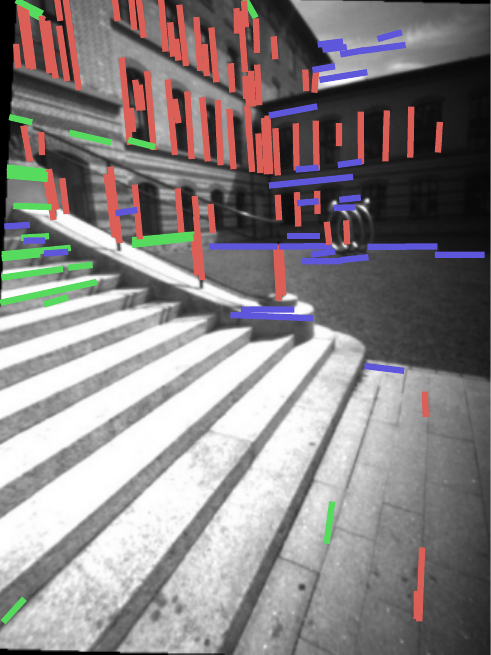}
        & \includegraphics[width=\ratiolamar\textwidth]{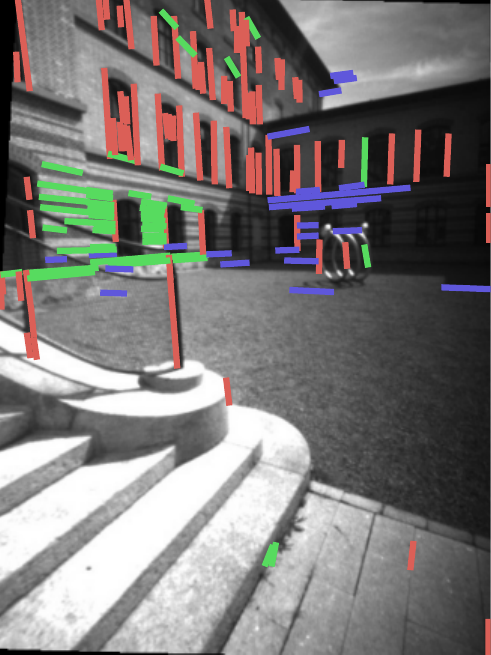} \\
        \multicolumn{2}{c}{Rot error = 17.98$^\circ$ ;  f error = 0.647} & \multicolumn{2}{c}{Rot error = 7.34$^\circ$ ; f error = 0.569} \vspace{0.1cm}\\
        \includegraphics[width=\ratiolamar\textwidth]{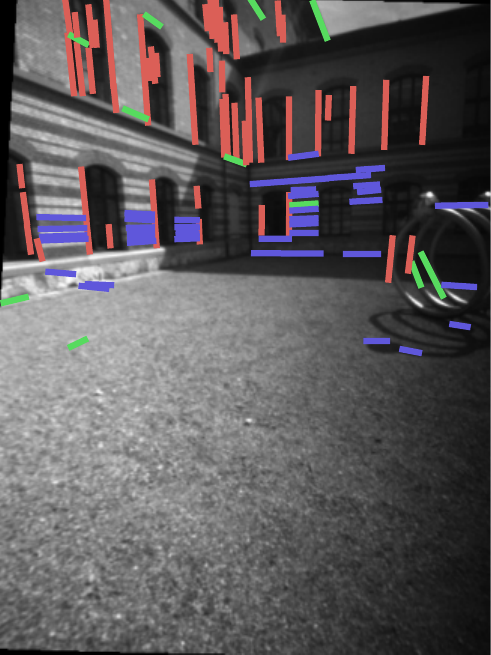}
        & \includegraphics[width=\ratiolamar\textwidth]{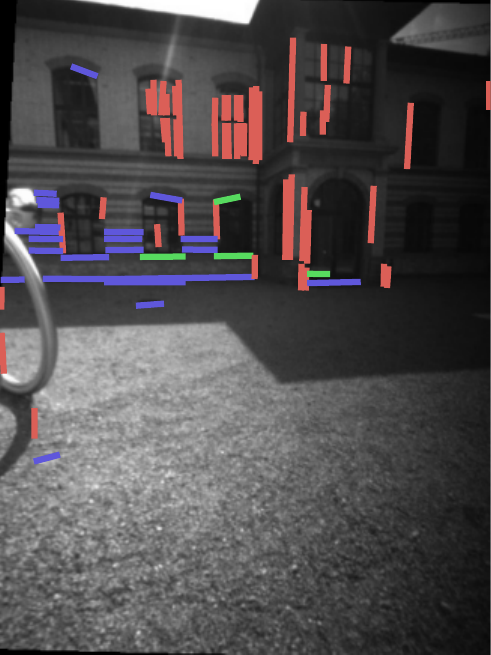}
        & \includegraphics[width=\ratiolamar\textwidth]{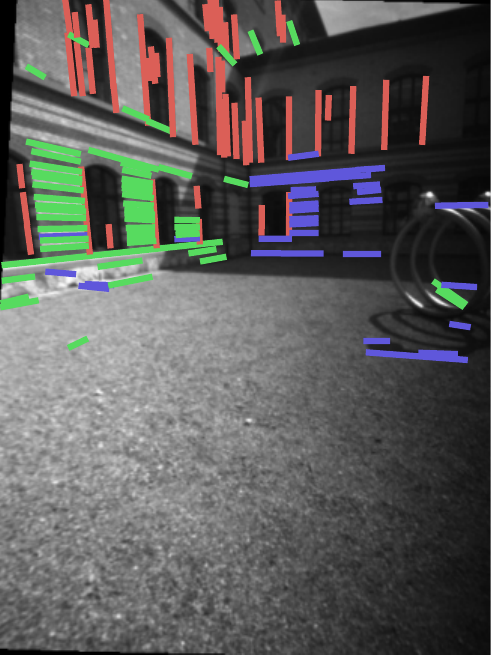}
        & \includegraphics[width=\ratiolamar\textwidth]{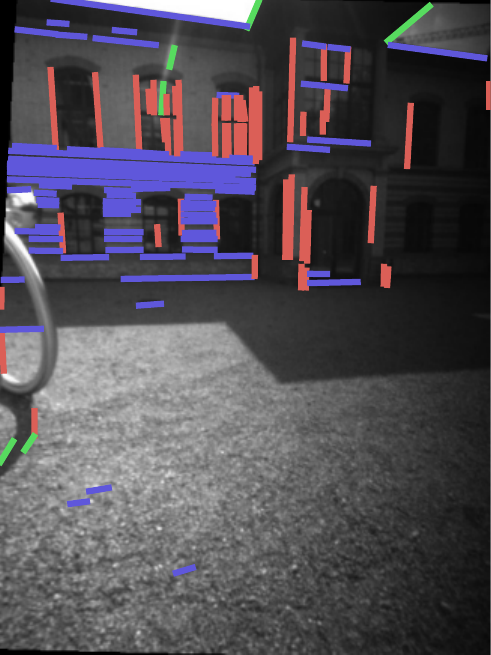} \\
        \multicolumn{2}{c}{Rot error = 27.82$^\circ$ ;  f error = 1.294} & \multicolumn{2}{c}{Rot error = 10.82$^\circ$ ; f error = 0.362} \vspace{0.1cm}\\
    \end{tabular}
    \caption{\textbf{Visualization of applications.} We display the inlier lines with one color per vanishing point, for the 2-1-1 solver of~\cite{wildenauer2012} (left), and our hybrid solver (right). The first three rows display pairs of images from an autonomous driving scenario on the KITTI dataset~\cite{Geiger2012CVPR}. Assuming that the gravity is vertical already gives a very good prior to our solver and boost the performance on relative rotation estimation. The last two rows are from an augmented reality setup on the LaMAR dataset~\cite{sarlin2022lamar}. Head movements are often purely rotational, and the IMU information is crucial to obtain accurate pose estimates.}
    \label{fig:visualizations_applications}
\end{figure*}

Additionally, to better highlight the application scenarios of our method, we show in Figure~\ref{fig:visualizations_applications} two cases where a prior on the gravity is available. The first one (first three rows) is for autonomous driving, where the cameras are usually always upright and the gravity can be assumed to be vertical. In the second example (last two rows), augmented and mixed reality (AR/MR) devices have usually an onboard IMU providing the gravity. In both cases, cars and AR/MR headsets are devices used over long periods of time, and the calibration of theirs cameras are subject to drift, and may need to be recalibrated on-the-fly. Thus, our method can benefit from the prior gravity available in both situations, while providing an estimate of the focal length at test time.

{\small
\bibliographystyle{ieee_fullname}
\bibliography{egbib}
}

\end{document}